\documentclass[table]{hfstyle/hf}

\usepackage{microtype}
\usepackage{hyperref}
\usepackage{url}
\usepackage{booktabs}
\usepackage{graphicx}
\usepackage{lineno}
\usepackage{enumitem}
\usepackage{listings} %
\usepackage{svg}

\definecolor{ darkblue}{rgb}{0, 0, 0.5}
\hypersetup{colorlinks=true, citecolor=darkblue, linkcolor=darkblue, urlcolor=darkblue}

\usepackage{amssymb}
\usepackage{multirow}
\usepackage{bigdelim}
\usepackage{todonotes}
\usepackage{longtable}
\usepackage{tabularray}
\usepackage{wrapfig}
\usepackage[most]{tcolorbox} 
\usepackage{url}
\usepackage{xspace}
\usepackage{svg}
\usepackage[absolute]{textpos} %

\usepackage{fdsymbol}   %
\usepackage{algorithm}
\usepackage{algpseudocode}

\usepackage[utf8]{inputenc} %
\usepackage[T1]{fontenc}    %
\usepackage{url}            %
\usepackage{booktabs}       %
\usepackage{amsfonts}       %
\usepackage{nicefrac}       %
\usepackage{microtype}      %
\usepackage[table]{xcolor}         %
\usepackage{amsmath}
\usepackage[most]{tcolorbox}
\usepackage{csquotes}

\usepackage{siunitx}
\usepackage{graphicx}
\usepackage{arydshln}
\usepackage{enumitem}
\usepackage{soul} %

\usepackage{multirow}
\usepackage{xspace}
\usepackage{adjustbox}
\usepackage{pifont}
\usepackage{caption}
\usepackage{makecell}
\usepackage{bold-extra}
\usepackage{url}

\usepackage{pgf-pie}

\usepackage{hyperref}
\definecolor{linkcolor}{RGB}{0, 0, 128}
\hypersetup{
     colorlinks   = true,
     citecolor    = linkcolor,
     linkcolor    = linkcolor,
     urlcolor     = linkcolor,
}
\usepackage{pifont}%
\usepackage{listings}

\setlist[itemize]{leftmargin=*,itemsep=0em,parsep=0.3em,topsep=0.3em}

\DeclareUnicodeCharacter{2212}{\ensuremath{-}}

\definecolor{maroon}{HTML}{F26035}
\definecolor{yellow}{HTML}{FDBC42}
\definecolor{lavender}{HTML}{734f96}
\definecolor{darkergrey}{HTML}{444444}
\definecolor{midgrey}{HTML}{e6eded}

\definecolor{neutralEight}{HTML}{343434}
\definecolor{neutralFive}{HTML}{838383}
\definecolor{neutralThree}{HTML}{bebebe}
\definecolor{neutralOne}{HTML}{dedede}
\definecolor{lightgrey}{HTML}{fafcfc}

\usepackage{tikz}

\definecolor{maroon}{HTML}{F26035}
\definecolor{yellow}{HTML}{FDBC42}
\definecolor{darkred}{RGB}{156, 39, 33}
\definecolor{darkblue}{RGB}{31, 90, 153}
\definecolor{forestgreen}{rgb}{0.13, 0.55, 0.13}
\definecolor{olmoDarkBlue}{HTML}{012e59}
\definecolor{olmoBlue}{HTML}{265ed4}
\definecolor{olmoLightBlue}{HTML}{012e59}
\definecolor{olmoTeal}{HTML}{00d5ff}
\definecolor{olmoYellow}{HTML}{ffbb00}
\definecolor{olmoOrange}{HTML}{ff9100}

\usepackage{setspace}

\usepackage{nicematrix}
\newcolumntype{L}[1]{>{\raggedright\let\newline\\\arraybackslash\hspace{0pt}}m{#1}}
\newcolumntype{C}[1]{>{\centering\let\newline\\\arraybackslash\hspace{0pt}}m{#1}}
\newcolumntype{R}[1]{>{\raggedleft\let\newline\\\arraybackslash\hspace{0pt}}m{#1}}
\newcolumntype{P}[1]{>{\centering\let\newline\\\arraybackslash\columncolor{ai2lightpink}}m{#1}}
\definecolor{lightyellow}{rgb}{1, 0.95, 0.85}
\definecolor{graphicbackground}{rgb}{0.9765,0.9451,0.9059}
\definecolor{codebackground}{rgb}{0.8314,0.949,0.9882}

\usepackage{booktabs}
\usepackage{graphicx}
\usepackage{subcaption}
\usepackage{multirow}

\newlength\savewidth

\newlength\thinwidth

\definecolor{Gray}{gray}{0.92}
\definecolor{DarkGray}{gray}{0.5}

\definecolor{LightCyan}{rgb}{0.88,1,1}
\definecolor{altRowColor}{gray}{0.92}
\definecolor{highlightRowColor}{rgb}{0.9, 0.9, 1}

\newcommand{\ours}{{SmolVLA}\xspace}

\title{
SmolVLA: A vision-language-action model for affordable and efficient robotics 
}

\newcommand{\sorbonne}{\raisebox{.15em}{\hspace{.08em}\includegraphics[height=.75em]{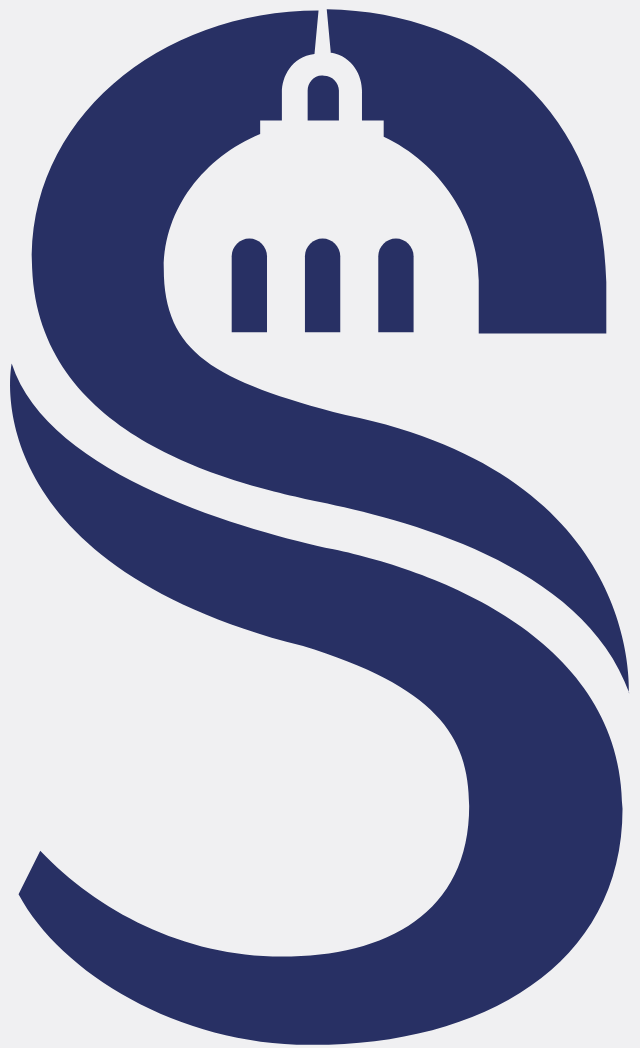}}\xspace}
\newcommand{\hf}{\raisebox{.28em}{\hspace{.05em}\includegraphics[height=.65em]{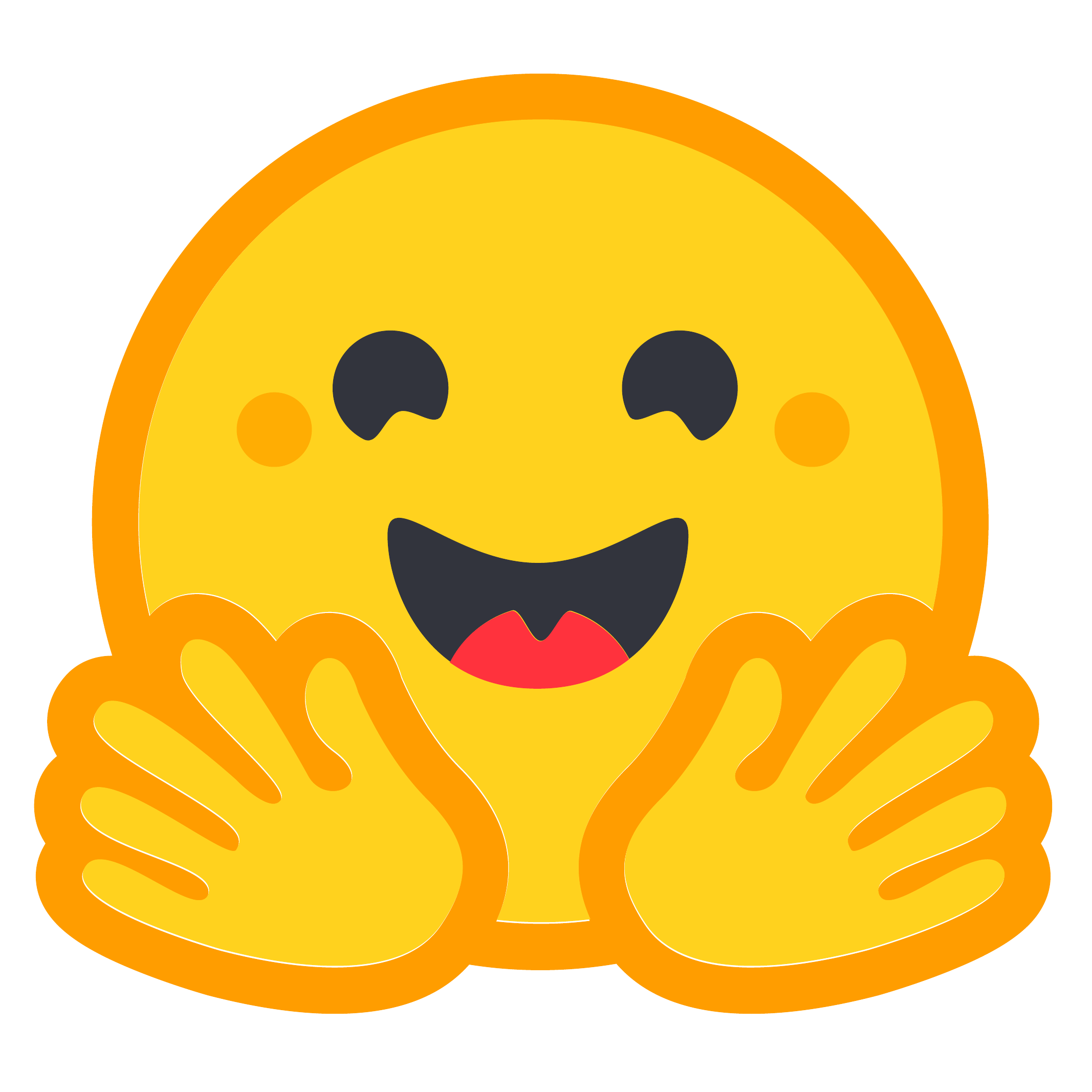}}\xspace}
\newcommand{\ensps}{\raisebox{.3em}{\hspace{.05em}\includegraphics[height=.65em]{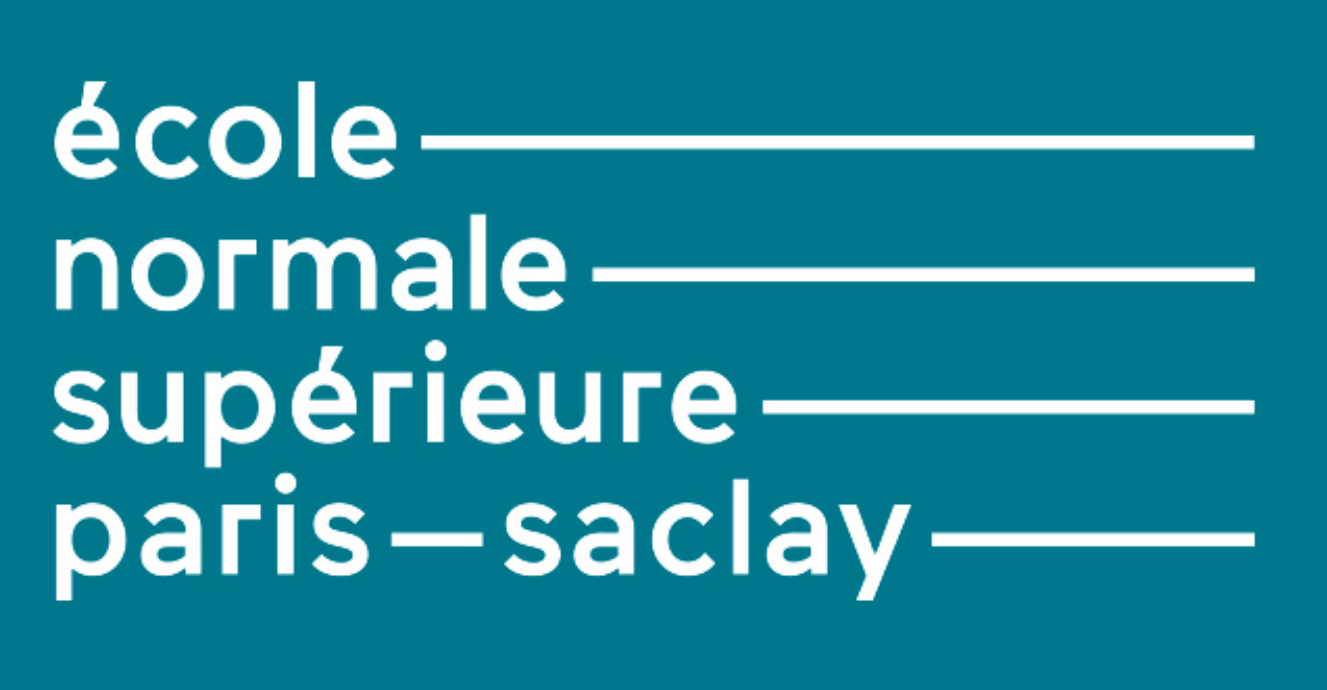}}\xspace}

\authorOne[]{Mustafa Shukor$^*$\sorbonne}
\authorOne[]{Dana Aubakirova$^*$\hf}
\authorOne[]{Francesco Capuano$^*$\hf\ensps}

\authorTwo[]{Pepijn Kooijmans\hf}
\authorTwo[]{Steven Palma\hf}
\authorTwo[]{Adil Zouitine\hf}
\authorTwo[]{Michel Aractingi\hf}
\authorTwo[]{Caroline Pascal\hf}
\authorTwo[]{Martino Russi\hf}
\authorTwo[]{Andres Marafioti\hf}
\authorTwo[]{Simon Alibert\hf}

\authorTwo[]{Matthieu Cord\sorbonne$^v$}
\authorTwo[]{Thomas Wolf\hf}
\authorTwo[]{Remi Cadene$^*$\hf}

\contribution[]{\hf Hugging Face, \sorbonne Sorbonne University, $^v$ valeo.ai, \ensps École Normale Supérieure Paris-Saclay \quad $^*$ Core team}

\abstract{
Vision-language models (VLMs) pretrained on large-scale multimodal datasets encode rich visual and linguistic knowledge, making them a strong foundation for robotics. 
Rather than training robotic policies from scratch, recent approaches adapt VLMs into vision-language-action (VLA) models that enable natural language-driven perception and control. 
However, existing VLAs are typically massive--often with billions of parameters--leading to high training costs and limited real-world deployability. 
Moreover, they rely on academic and industrial datasets, overlooking the growing availability of community-collected data from affordable robotic platforms.
In this work, we present SmolVLA, a small, efficient, and community-driven VLA that drastically reduces both training and inference costs, while retaining competitive performance. 
SmolVLA is designed to be trained on a single GPU and deployed on consumer-grade GPUs or even CPUs. 
To further improve responsiveness, we introduce an asynchronous inference stack decoupling perception and action prediction from action execution, allowing higher control rates with chunked action generation. 
Despite its compact size, SmolVLA achieves performance comparable to VLAs that are 10$\times$ larger. We evaluate SmolVLA on a range of both simulated as well as real-world robotic benchmarks and release all code, pretrained models, and training data.

}

\begin{document}

\maketitle

\begin{figure}[h]
    \centering
    \begin{minipage}[t]{1\textwidth}
        \centering
        \includegraphics[width=0.88\textwidth]{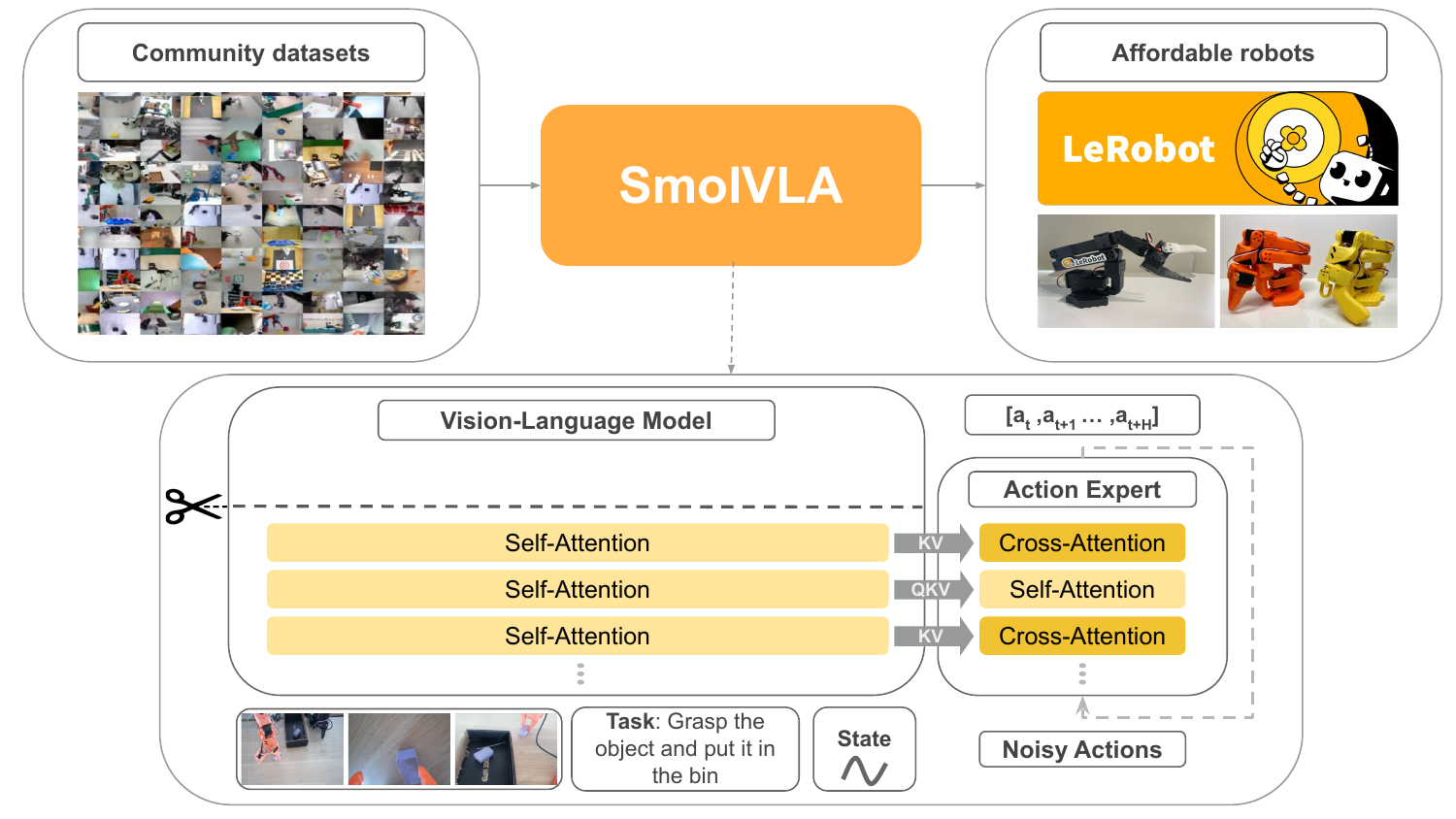}
        \caption{\footnotesize \textbf{SmolVLA.} SmolVLA consists of a compact pretrained vision-language model, discarding the last $L-N$ layers (scissors icon). The remaining layers embed three inputs: (i) language instruction, (ii) RGB image(s), and (iii) robot sensorimotor state. Their merged tokens feed an Action Expert of alternating cross-attention (gold) and self-attention (light yellow) blocks, trained with flow matching to output $n$ low-level actions chunk $a_t, \dots, a_{t+n}$. SmolVLA is pretrained on public community datasets and evaluated on low-cost robots.
        }
        \label{fig:arch}
    \end{minipage}
    \vspace{-0.6cm}
\end{figure}

\section{Introduction}

In recent years, the field has shifted towards the development of \emph{foundation} models, generalist models capable of performing a wide range of tasks.
A prominent example of this trend are large language models (LLMs), which have demonstrated performance comparable to the average human in understanding and generating natural language, reasoning over complex topics, and anchoring in knowledge~\citep{GPT3,GPT4,Llama3,Gemini,Mistral7B}. 
The success of text-based models has thus been extended to other modalities, sparking interest towards multi-modal vision-language (VLMs)~\citep{alayrac2022flamingo,PaLI-3,kosmoshuang2023language,LLaVA-1.5,chen2024internvl25,shukor2023unival} and audio-language models (ALMs)~\citep{defossez2024moshi,das2024speechverse,borsos2023audiolm}. 
While complementary in terms of modalities, such advances in developing multi-modal foundation models stem from \emph{(i)} the adoption of scalable architectures, such as the Transformer~\citep{vaswani2017attention} and \emph{(ii)} internet-scale training datasets.

Despite their remarkable achievements in the digital world, real-world application of foundation models--particularly in robotics--remains limited. 
In particular, robotic policies~\citep{zhao2023learningact,chi2024diffusionpolicy,lee2024behavior,Hansen2022tdmpc} still face challenges in generalizing across object types, positions, environments, and tasks~\citep{xie2024decomposing,ebert2021bridge}. 
Robots should be able to adapt to new surroundings and new objects, which requires robust skills and common sense understanding of the world. Yet, the progress in this direction seems to be often limited by the availability of high-quality and diverse data.

To address this limitation, a growing body of work has begun exploring robotics foundation models in the form of vision-language-action (VLA) models~\citep{team2024octo,o2024openrtx,brohan2023rt2,kimopenvla,black2024pi_0,bjorck2025gr00t,li2024vision,huang2024embodied}.
VLAs are designed to incorporate abstract reasoning, world knowledge, and decision-making skills embedded in pretrained large language and vision-language models. 
These models take multimodal inputs--such as visual observations and natural language instructions--and predict the corresponding robotic actions. 
Early results suggest promising gains in generalization capabilities \citep{black2024pi_0,brohan2023rt2}.

VLA models remain in an early stage of development and are not yet as mature or widely adopted as LLMs and VLMs. Much of the impactful VLA progress remains proprietary, with many models sharing only weights while withholding full training details and essential methodological components.
While effective in tackling academic benchmarks, we argue achieving human-level capabilities in robotics requires a stronger commitment to open-source efforts. In particular, transparent, reproducible open-source models and training recipes are crucial for accelerating progress and fostering broader participation within the robotics research community.
We advocate for the development of affordable, efficient models that are accessible to a broader community. 
While encouraging efforts like OpenVLA \citep{kimopenvla} and RT-2-X \citep{o2024openrtx} demonstrate the feasibility of open VLA systems, they remain large, resource-intensive, and dependent on costly robotic platforms, hindering accessibility.

In this work, we introduce \ours, an open-source initiative featuring a compact yet capable VLA model, released alongside reproducible and efficient training and inference recipes.
Our contributions are as follows:
\begin{itemize}
    \item \textbf{Lightweight architecture.} We present \ours, a compact and efficient vision-language agent optimized for training on consumer-grade GPUs and deployment on CPUs. Key design choices include: \emph{(i)} skipping layers in the VLM, \emph{(ii)} using a minimal number of visual tokens \emph{(iii)} leveraging small pretrained VLMs, and \emph{(iv)} interleaving self-attention layers with lighter cross-attention layers.
    \item \textbf{Pretraining on community-driven datasets.} \ours is trained end-to-end on fewer than 30k episodes drawn entirely from publicly available, community-contributed datasets, demonstrating strong performance while using an order of magnitude less data than prior art.
    \item \textbf{Asynchronous inference.} We introduce an optimized asynchronous inference stack that decouples action execution from observation processing and action prediction, reducing latency and enabling fast, resource-efficient inference.
\end{itemize}

We evaluate \ours on both simulated environments and real-world settings on multiple tasks. Interestingly, although significantly smaller, \ours matches or surpasses the performance of much larger VLA models.

\section{Related work}

\paragraph{Vision-language models (VLMs).} VLMs are designed to process both visual and textual modalities--most commonly by taking both images and text as input and generating text conditioned on the visual context. 
Recent advances in VLMs have been driven by the success of LLMs, with many approaches building upon pretrained LLMs and adopting similar training paradigms. 
Typically, VLMs~\citep{alayrac2022flamingo,laurenccon2024mattersidefics2,VILA} are constructed by integrating a pretrained vision encoder~\citep{radford2021learning,li2023siglip,fini2024multimodalaimv2} with a pretrained LLM~\citep{llama3_2modelcard,Mistral7B,wang2024qwen2}. 
Training then proceeds in multiple multimodal stages, beginning with large-scale pretraining on image-caption datasets~\citep{LAION-COCO,COYO-700M} and interleaved vision-language corpora~\citep{OBELICS,MMC4}, all followed by a supervised fine-tuning stage on instruction-tuning datasets~\citep{LLaVA-1.5,tong2024cambrian,laurenccon2024mattersidefics2}. 
Other works have shown the benefits of not relying on pretrained vision encoders~\citep{fuyu-8b,shukor2025scaling,diao2025evev2,diao2024unveiling}, while other works aims at developing more unified architectures representing both images and text as discrete tokens, enabling a single model to process multimodal sequences of tokens~\citep{wang2022ofa,shukor2023unival,team2024chameleon,lin2024moma}. 
Efficiency has also become a central focus in VLM research. Several works aim to reduce training costs by using smaller, more diverse datasets~\citep{LLaVA-1.5,InstructBLIP,bai2025qwen25vl,zhu2024minigpt,tong2024cambrian}, training smaller-scale models~\citep{marafioti2025smolvlm,moondream,minicmpv2024}, or by adapting pretrained unimodal models by tuning only a small subset of parameters~\citep{shukor2023epalm,vallaeys2024improveddepalm,MAPL,FROMAGe,tsimpoukelli2021multimodalfrozen,BLIP-2}. 
While the majority of VLM research focuses on image and text modalities, recent work has demonstrated that similar techniques can be extended to integrate additional modalities, such as video and audio~\citep{wang2025internvideo2,liu2024kangaroo,zhang2025videollama,kong2024audioflam}.

\paragraph{Vision-language-action models (VLAs).} A growing area of interest in robotics research is the development of generalist policies--models capable of performing a wide range of tasks, generalizing across different environments and robot embodiments. 
A prominent strategy in this direction leverages VLAs, models capable of processing \emph{(i)} task instructions given in natural language, \emph{(ii)} visual observations (e.g., images coming from camera streams), and \emph{(iii)} proprioceptive inputs to output control actions.
Early efforts such as Octo~\citep{team2024octo} and RT-1~\citep{o2024openrtx} trained transformer-based models from scratch on large-scale robotic demonstration datasets. To improve both performance and generalization, RT-2~\citep{brohan2023rt2} leveraged pretrained vision-language models (VLMs), further training them on robotics-specific data. 
In an effort to promote openness and reproducibility, OpenVLA~\citep{kimopenvla} released a 7B-parameter VLA trained on publicly available data to generate discrete action tokens. 
As action tokenization poses limitations for continuous control, \( \pi_0 \) \citep{black2024pi_0} and DexVLA \citep{wen2025dexvla} proposed using diffusion-based decoders for continuous action generation. 
In this, both~\citet{black2024pi_0, wen2025dexvla} propose adapting a pretrained VLM, RDT-1B,introducing a large diffusion component--termed \emph{action expert}--trained directly on robotic demonstrations. 
Recently,~\cite{pertsch2025fast} proposed a fully autoregressive approach using a novel action tokenizer, improving over traditional binning methods but still suffering from slow (autoregressive) inference. 
In an effort to improve VLAs' efficiency, TinyVLA~\citep{wen2024tinyvla} trained a lightweight sub-1B model from scratch on multimodal data and then fine-tuned it on robotics datasets,although the absence of large-scale pretraining on robotics data hinders wider generalization capabilities. 
\ours shares similar goals with most of these efforts, aiming to develop and release open-source models that are both performant and highly efficient in terms of training and inference.

\newcommand{\actionchunk}{\mathbf{A}}
\newcommand{\actionexpert}{\mathbf{v}_\theta}

\section{SmolVLA: small, efficient and capable}

\paragraph{Overview.}
\ours is a lightweight VLA composed of a compact pretrained VLM, and an action expert trained with flow matching. 
Given multiple images and a language instruction describing the task, the model outputs a chunk of actions. 
It is first pretrained with imitation learning on community-collected datasets, then evaluated across both real-world and simulated environments. 
The pretraining data is designed to span a diverse set of tasks and behaviors, allowing the model to learn generalizable physical skills that transfer across settings. 
At inference time, we introduce an asynchronous execution stack that decouples action execution from perception and prediction, enabling faster and more responsive control.

\subsection{Model architecture}
SmolVLA is composed of two main components: \emph{(i)} a pretrained VLM tasked with perception and \emph{(ii)} an action expert, trained to act. 
The two components are interconnected, as the VLM processes state inputs to generate features that condition the action expert, which produces actions in turn altering the state fed to the VLM. 
Specifically, the VLM processes sensorimotor states, including images from multiple RGB cameras, and a language instruction describing the task. In turn, the VLM outputs features directly fed to the action expert, which outputs the final continuous actions.

\paragraph{Vision-language model (VLM).}
We leverage a pretrained VLM as the main backbone for perceiving the robot’s environment. VLMs, pretrained on diverse multimodal data, capture rich world knowledge. To guarantee efficiency and accessibility, we choose SmolVLM-2~\citep{marafioti2025smolvlm}, an efficient model optimized for multi-image and video inputs. 
SmolVLM-2 relies on SigLIP~\citep{li2023siglip} to encode visual features for SmolLM2 language decoder~\citep{allal2025smollm2}.
In \ours, the VLM component processes image sequences using the vision encoder, which reduces token count through a token-shuffling technique for efficiency. The language instruction is tokenized into text tokens. Sensorimotor states are projected into a single token via a linear layer to match the language model’s token dimension. Lastly, visual, language, and state tokens are concatenated and passed to the language decoder. The resulting obtained via the decoder layers are then used to condition the action expert.

\paragraph{State, action, and feature projectors.}
We use linear projection layers in various points inside of \ours. In particular, we use linear projection layers to \emph{(i)} project the states to match the VLM dimension \emph{(ii)} project the actions to match the action expert dimensions and \emph{(iii)} to adapt the VLM features to align with the action expert’s dimension.

\paragraph{Visual tokens reduction.}
While proven critical for VLM performance, high-resolution images increase inference costs. 
To guarantee efficiency, SmolVLM-2 is trained with image tiling~\citep{lin2023sphinx}, a popular technique that involves processing multiple crops of the same image, in addition to the global image. 
However, to obtain a faster inference time, we do not use tiling. We use the global image only, in addition to a pixel shuffle operation, limiting the visual tokens to 64 per frame.

\paragraph{Faster inference through layer skipping.}
To get faster inference time, we skip computations in the VLM. Prior work~\citep{shukor2024skipping,tang2023you} demonstrated the possibility of skipping layers in pretrained models without incurring in significant performance degradation. 
Recently,~\citep{el2024scalable,bolya2025perception,rajasegaran2025empirical} have shown that the best features for downstream tasks are not necessarily obtained from the last layer of a VLM. 
Consequently, rather than using the last layer features, our action expert has access to all features up to a specified layer \( N \). 
In practice, we find setting \( N \) to half the total layers (\( N=L/2 \)) offers a good tradeoff between speed and performance, effectively halving the LLM and action expert’s computational cost.

\paragraph{Flow matching action expert.}
The action expert \( \mathbf{v}_{\theta} \) is trained to predict an action chunk \( \actionchunk_t = ( a_t, \ldots, a_{t+n} ) \) from VLM features.
In keeping with prior work, our implementation of \( \mathbf{v}_\theta \) relies on the transformer architecture \citep{vaswani2017attention}.
Differently from prior VLA architectures, we interleave cross and self-attention layers, thus using a conditional Flow Matching Transformer~\citep{esser2024scaling,liu2022rectified,lipman2022flow} as \(  \mathbf{v}_{\theta} \). 
The action expert is trained using the objective defined by:

\begin{equation*}
\mathcal L ^\tau(\theta) = \mathbb{E}_{p(\actionchunk_t \mid \mathbf{o}_t), q(\actionchunk_t^\tau \mid \actionchunk_t)} \left[ \left\| \mathbf{v}_{\theta}(\actionchunk_t^\tau, \mathbf{o}_t) - \mathbf{u}(\actionchunk_t^\tau \mid \actionchunk_t) \right\|^2 \right],
\end{equation*}

where $\mathbf{o}_t$ represents the VLM features extracted from an observation \( o_t \) at the \( N \)-th VLM layer, and \( \actionchunk_t^\tau = \tau \actionchunk_t + (1 - \tau) \epsilon \), with \( \epsilon \sim \mathcal{N}(0, \mathbf{I}) \). 
In particular, \( \mathbf{v}_{\theta} \) is trained to output the vector field \( \mathbf{u}(\actionchunk_t^\tau \mid \actionchunk_t) = \epsilon - \actionchunk_t \) from the VLM features and the noisy actions \( \actionchunk_t^\tau \). 
In keeping with~\citet{black2024pi_0}, we sample \( \tau \) from a Beta distribution. To improve on inference's efficiency, we use a reduced hidden size of \( 0.75 \times d \) for \( \mathbf{v}_\theta \), where \( d \) is the VLM's hidden dimension.

\paragraph{Interleaved cross and causal self-attention layers.}
The action expert \( \actionexpert \) generates action chunks conditioned on the VLM features, with the interaction between the VLM and action expert in \ours being facilitated by the attention mechanisms.
Unlike prior works relying exclusively on self-attention (SA)~\citep{black2024pi_0} or cross-attention (CA)~\citep{bjorck2025gr00t}, we employ an interleaved approach, where each block contains either a CA or a SA layer. 
This design choice also differs from standard VLM architectures, where each decoder block typically includes both SA and CA layers~\citep{OBELICS,alayrac2022flamingo,chen2022pali}. 
Within the action expert's forward pass, the interaction between actions and VLM features takes place via attention, projecting tokens into query, keys and values~\citep{vaswani2017attention}. 
In our setup, CA layers cross-attend the VLM's keys and values, while SA layers allow the action tokens in \( \actionexpert \) to attend to each other.
We employ a causal attention mask for the SA layers, ensuring that each action token can only attend to past tokens within the chunk, preventing future action dependencies. Empirically, we find that interleaving CA and SA layers provides higher success rates and faster inference time. In particular, we find self-attention to contribute to smoother action chunks \( \actionchunk \), something particularly evident when evaluating on real robots.

\subsection{Pretraining data collected by the community}

In robotics, the data available for large-scale pretraining remains orders of magnitude smaller than what has driven recent breakthroughs in vision and language. 
For instance, while natural language foundation models can benefit from the unique text-based interface and massive scale of internet-data, the integration and scaling up of robotics datasets appears complex, due to the \emph{(i)} differences across datasets and \emph{(ii)} reliance on teleoperation by human experts for data collection.
Additionally, the high heterogeneity of robot morphologies, sensors, actuation modes, control frequencies, and data formats results in "data islands"~\cite{bjorck2025gr00t}--scattered robotics datasets whose integration proves challenging.

In this context, the advent of low-end robotic platforms and standardized robotics libraries directly mitigates this data heterogeneity, providing a unique entry point to robotics for practitioners. 
Further, open-source data contributions collected by individual practitioners enable the larger robotics community with \textit{community datasets}, datasets collected in diverse real-world settings--from academic labs to homes--as part of a larger effort to decentralize and scale robot learning with open-source.
Unlike academic datasets following standardized protocols, community datasets naturally span varied robot embodiments, control schemes, camera perspectives, and tasks. Further, community datasets reflect real-world complexity through noisy demonstrations, heterogeneous environments, and diverse object interactions, providing valuable as pre-training data. In this work, we selected a subset of 481 community datasets obtained from Hugging Face, filtered according to embodiment type, episode count, overall data quality, and frame coverage (Table~\ref{tab:pretraining_data}).

\vspace{-0.4cm}
\begin{wraptable}{r}{0.4\linewidth}
    \centering
    \setlength{\tabcolsep}{8pt} %
    \renewcommand{\arraystretch}{1} %
    \resizebox{\linewidth}{!}{
        \begin{tabular}{ccccccc}
            \toprule
            \textbf{\# datasets} & \textbf{\# episodes} & \textbf{\# frames}  \\
              481 & 22.9K & 10.6M\\
            \bottomrule
        \end{tabular}
    }
    \captionof{table}{\textbf{Community datasets statistics.} At \(\sim\) 10M episodes, our pretraining set stands at least one order of magnitude smaller than other state-of-the-art. For completeness, we report the list of community-datasets used in \Cref{app:comm_data}.}
    \vspace{-0.4cm}
    \label{tab:pretraining_data}
\end{wraptable}

\paragraph{Task annotation with VLM.} 
Relying on community-contributed datasets entails standardization challenges. In particular, we observed substantial noise in task annotations--natural language descriptions of the robot’s intended behavior for a given dataset. 
Critically, various datasets included ambiguous placeholders such as \texttt{task desc}, overly vague commands such as \texttt{Hold} or \texttt{Up}, or lacked instructions entirely. To improve on the annotation quality, we used an off-the-shelf VLM (\texttt{Qwen2.5-VL-3B-Instruct}) to auto-generate concise task descriptions. For each dataset, we sampled representative frames and provided them alongside the original instruction. The model was prompted to produce a short, action-oriented sentence summarizing the behavior. The full prompt is available in \Cref{app:comm_data}.

\paragraph{Camera viewpoint normalization.} 
Another challenge arising from using community datasets consists in the high variability in the camera naming conventions used. For instance, datasets refer \texttt{images.laptop} may refer to a top, side, or wrist-mounted view depending on the particular. We found this inconsistency detrimental during pretraining, and that a consistent camera ordering is rather beneficial for training in this data regime. To address this standardization challenge, we manually mapped each camera to a standardized view type—prioritizing top, wrist, and side perspectives--and renamed them as \texttt{OBS\_IMAGE\_1}, \texttt{OBS\_IMAGE\_2}, and \texttt{OBS\_IMAGE\_3}, respectively. 
For datasets with additional views, the order was preserved, but unused views were dropped during training. Future efforts may automate this process using VLMs, or propose/adopt standardized data collection guidelines.

\subsection{Asynchronous inference}\label{sec:async}

Modern visuomotor policies~\citep{zhao2023learningact, chi2023diffusion, black2024pi_0} output \emph{action chunks}--sequences \(\pi(o_t) = \actionchunk_t \) with \(\actionchunk_t = \bigl(a_t,a_{t+1},\dots,a_{t+n}\bigr) \) being a sequence of \(n \) (much greater than 1) low-level commands enqueued in an action queue, originating from an environment observation, \(o_t\).
Typically, the robot executes the entire action chunk \(\actionchunk_t \), before a new observation \( o_{t+n} \) is passed to the policy \( \pi \) to predict the next chunk. This results in open-loop inference in between observations captured every \( n \) timesteps.
Works including~\citet{zhao2023learningact, chi2023diffusion} adopt a different strategy whereby the robot controller interleaves chunk prediction \( \actionchunk_t \gets \pi(o_t) \) and chunk consumption \( a_t \gets \textsc{PopFront(\( \actionchunk_t \))} \), computing a new chunk of actions at every timestep \( t \) and aggregating the predicted chunks on overlapping sections.
While adaptive--every observation at every timestep \( o_t\) is processed--such approaches rely on running inference continuously, which can be prohibitive in resource-constrained scenarios, such as edge deployments.

A less resource-intensive approach is to entirely exhaust the chunk \( \actionchunk \) before predicting a new chunk of actions, a strategy we refer to as \textit{synchronous} (sync) inference. 
Moreover, sync inference efficiently allocates computation every \( n \) timestep, resulting in a reduced average computational burden at control time. In contrast, it inherently hinders the responsiveness of robot systems, introducing blind lags due to the robot being \textit{idle} while computing \( \actionchunk \).

\begin{figure}
    \centering
    \begin{minipage}[t]{\textwidth}
        \centering
        \includegraphics[width=0.9\textwidth]{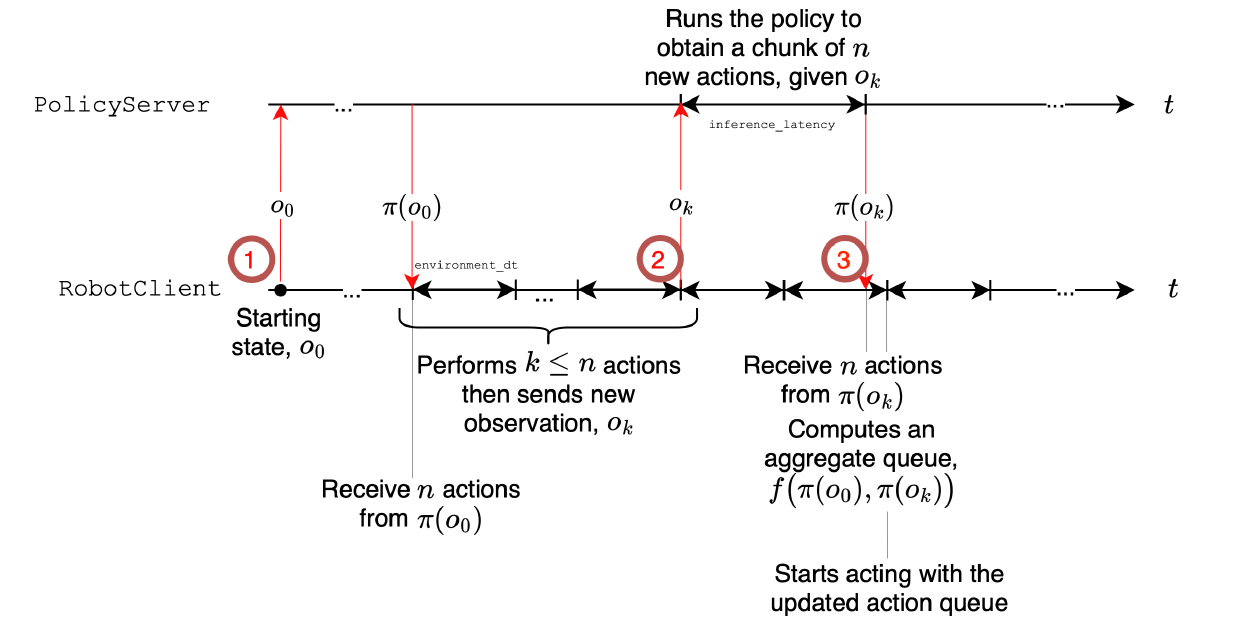}
        \caption{\textbf{Asynchronous inference}. Illustration of the asynchronous inference stack. Note that the policy can be run on a remote server, possibly with GPUs.}
        \label{fig:async-inference}
    \end{minipage}
    \vspace{-0.6cm}
\end{figure}

We directly assess the lack of adaptiveness of robot systems due to acting open-loop, and the presence of lags at runtime by decoupling action chunk prediction \( \actionchunk \) from action execution \( a_t \gets \textsc{PopFront}(\actionchunk_t) \), developing an \textit{asynchronous} (async) inference stack (\Cref{alg:async-inference}), whereby a \( \textsc{RobotClient} \) sends an observation \( o_t \) to a \( \textsc{PolicyServer} \), receiving an action chunk \( \actionchunk_t \) once inference is complete (\Cref{fig:async-inference}).
In this, we avoid execution lags by triggering chunk prediction while the control loop is still consuming a previously available queue, aggregating it with the newly incoming queue whenever available.
In turn, async-inference tightens the loop between action prediction and action execution, by increasing the frequency at which observations are processed for chunk prediction. 
Crucially, decoupling action prediction from action execution also directly allows to allocate more computational resources on a remote policy server sending actions to the robot client over networks, something which may prove very effective in resource-constrained scenarios such as low-power robots.

\begin{algorithm}
  \caption{Asynchronous inference control-loop}
  \label{alg:robotclient}
  \begin{algorithmic}[1]
    \State \textbf{Input:} horizon $T$, chunk size $n$, threshold $g\in[0,1]$
    \State \textbf{Init:} capture $o_0$; send $o_0$ to \textsc{PolicyServer};
           receive $\actionchunk_0 \gets \pi(o_0)$
    \For{$t$ \textbf{to} $T$}
        \State $a_t \gets \textsc{PopFront}(\actionchunk_t)$
        \State \textsc{Execute}($a_t$) \Comment{execute action at step $t$}
        \If{$\tfrac{|\actionchunk_t|}{n} < g $} \Comment{queue below threshold}
            \State capture new observation, $o_{t+1}$
            \If{\textsc{NeedsProcessing}$(o_{t+1})$} \Comment{similarity filter, or triggers direct processing}
                \State \texttt{async\_handle} $\gets \textsc{AsyncInfer}(o_{t+1})$ 
                \Comment{Trigger new chunk prediction (non blocking)}
                \State $\tilde{\actionchunk}_{t+1} \gets \pi(o_{t+1})$ \Comment{New queue is predicted with the policy}
                \State $\actionchunk_{t+1} \gets f(\actionchunk_t,\tilde{\actionchunk}_{t+1})$ \Comment{aggregate overlaps (if any)}
                
            \EndIf
        \EndIf
        \If {\textsc{NotCompleted}(\texttt{async\_handle})}
            \State $\actionchunk_{t+1} \gets \actionchunk_t$ \Comment{No update on queue (inference is not over just yet)}
        \EndIf
    \EndFor
  \end{algorithmic}
  \label{alg:async-inference}
\end{algorithm}

\paragraph{Implementation details}
\textit{Async} inference \emph{(i)} tightens the control loop by capturing observations more often, directly eliminates idle gaps at runtime, and \emph{(ii)} directly allows to run inference on more powerful computational resources than the ones typically available onboard autonomous robotic platforms.

Algorithmically, we attain \emph{(i)} on the \textsc{RobotClient}-side by consuming actions from a readily available queue until a threshold condition on the number of remaining actions in the queue (\(\vert \actionchunk_t \vert / n < g \)) is met. When this condition is triggered, a new observation of the environment is captured and sent to the (possibly remote) \textsc{PolicyServer}. 
To avoid redundant server calls and erratic behavior at runtime observations are compared in joint-space, and near-duplicates are dropped.
Two observations are considered near-duplicates if their distance in joint-space is under a predetermined threshold, \( \epsilon \in \mathbb R_+\).
Importantly, when the queue available to robot client eventually becomes empty, the most recent observation is processed regardless of similarity.

Interestingly, the behavior of async inference can be studied analytically. First, let \( \ell \) be a random variable modeling the time needed to receive an action chunk \( \actionchunk \) after sending an observation \( o \), i.e. the sum of \emph{(i)} the time to send across the observation \( o \) between the \textsc{RobotClient} and \textsc{PolicyServer}, \( t_{C \to S}\) \emph{(ii)} the inference latency on the \textsc{PolicyServer}, \( \ell_S \) and \emph{(iii)} the time to send \( \actionchunk \) between the \textsc{PolicyServer} and \textsc{RobotClient}, \( t_{S \to C} \). Assuming independence, \( \mathbb E [\ell] = \mathbb E[t_{C \to S}] + \mathbb E[\ell_S] + \mathbb E[t_{S \to C}] \) which can be further simplified to \( \mathbb E[\ell] \simeq \mathbb E[\ell_S]  \), assuming communication time is \emph{(i)} equal in both directions and \emph{(ii)} negligible with respect to the inference latency. Second, let \(\Delta t\) be the environment’s control cycle. With a real-world frame-rate of 30 frames per second, \(\Delta t=33\text{ms}\). Consequently, exhausted queues at runtime--i.e. being idle awaiting for a new chunk--are avoided for \( g \geq \frac{\mathbb E[\ell_S] / \Delta t}{n} \). In this, the queue threshold \( g \) plays a major role relatively to the availability of actions to the \textsc{RobotClient}.

\begin{figure}
    \centering
    \begin{minipage}[t]{0.99\textwidth}
        \centering
        \includegraphics[width=\textwidth]{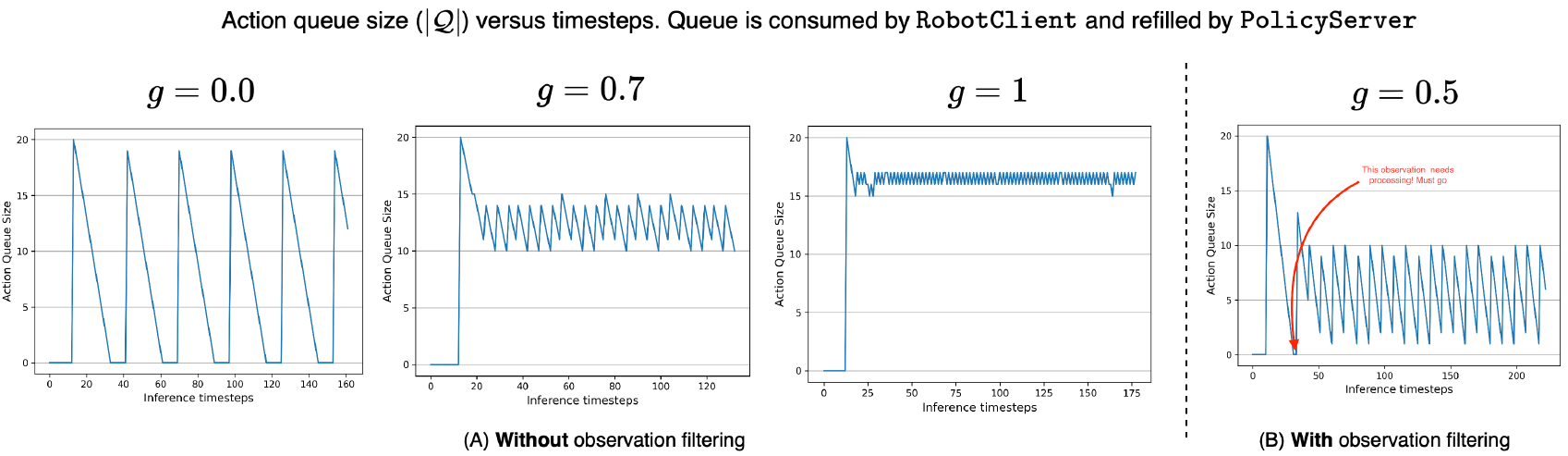}
        \caption{Action queue size evolution at runtime for various levels of \( g\) when (A) not filtering out observation based on joint-space similarity and (B) filtering out near-duplicates observation, measuring their similarity in joint-space.}
        \label{fig:queues}
    \end{minipage}
\end{figure}

\Cref{fig:queues}(A) illustrates how the size of the action chunk \(\lvert \actionchunk_t \rvert\) evolves over time for three representative values of \(g\), detailing the following key scenarios:
\begin{itemize}
    \item \textbf{Sequential limit \((g=0)\).} The client drains the entire chunk before forwarding a new observation to the server. During the round-trip latency needed to compute the next chunk, the queue is empty, leaving the robot \emph{incapable of acting}.  This reproduces the behavior of a fully sequential deployment and results in an average of \( \mathbb E[\ell_S] \) idle seconds.
    \item \textbf{Asynchronous inference \((g=0.7)\).} Allowing the client to consume a fraction of roughly \(1-g = 0.3\) of a queue \( \actionchunk_{t-1}\) before triggering inference for a new action queue \( \actionchunk_{t} \), amortizing computation while keeping the queue from emptying. The overlap between successive chunks provides a buffer against modeling errors without the full cost of the \(g=1\) regime. The updated queue \( \actionchunk_t\) is obtained aggregating queues on the overlapping timesteps between \( \actionchunk_{t-1}\) and the incoming \(\tilde{\actionchunk}_{t}\).
    \item \textbf{Compute-intensive limit \((g=1)\).}  As an extreme case, and in keeping with \citet{zhao2023learningact, chi2024diffusionpolicy}, an observation is sent at \emph{every} timestep. The queue is therefore almost always filled, with only a minor saw-tooth due to \(\Delta t/\mathbb E[\ell_s] < 1\). While maximally reactive, this setting incurs one forward pass per control tick and can prove prohibitively expensive on limited hardware. Importantly, because the client is consuming actions while the server computes the next chunk, the available queue never gets filled again.
\end{itemize}

\Cref{fig:queues}(A) emphasizes the trade-off governed by \(g\): small values place result in idle periods, whereas \(g\approx 1\) assumes a highly accurate model and pays a significant compute price. In practice, choosing \(g\in(0,1)\) allows to strike a balance between reactivity against resource budgets. 
If not for the aforementioned similarity filter, the \textsc{RobotClient} would send observations for processing every \( (1 - g) n \cdot \Delta t\) seconds, receiving a new chunk of actions every \( (1 - g) n \cdot \Delta t + \mathbb E[\ell_S] \), on average. 
The presence of the observation similarity filter dilates this processing time, and serves the scope of avoiding the robot stalling due to the queue being constantly integrated with an incoming, nearly identical, action chunk. 
In particular, ~\Cref{fig:queues}(B) results in a queue which is filled with incoming actions \emph{unless} near-duplicate observations are filtered out from the processing pipeline. For clarity, the red arrow in \Cref{fig:queues}(B) highlights a timestep where the observation similarity mechanism is bypassed, forcing a (nearly identical) observation to be processed as the queue results empty.

\section{Experiments}

\subsection{Experimental setup}
We evaluate our model on both simulated and real-world robotic manipulation tasks. 
To evaluate \ours in simulation, we collected a new dataset for MetaWorld~\citep{yu2020metaworld} comprising of 50 demonstrations for each of the 50 tasks.
For real-world evaluation, we collected three datasets using the SO-100 robot arm~\citep{Knight_Standard_Open_SO-100} and 1 with SO-101 arm~\citep{Knight_Standard_Open_SO-100}, each corresponding to a different manipulation task. Each dataset contains demonstrations relative to one task, with 10 trajectories for each of 5 distinct starting positions, resulting in a total of 50 demonstrations per dataset. The datasets record trajectories relative to
Unless specified otherwise, \ours is always trained in a multi-task setting.

\paragraph{Evaluation metrics.} We report success rate (SR) as the primary metric across all benchmarks. For simulation-based evaluations, SR is binary--set to 1 if the task is successfully completed, and 0 otherwise. For real-world evaluations, we adopt a more fine-grained scoring approach by decomposing each task into subtasks. For example, in the Pick-and-Place task, we assign a score of 0.5 for successfully picking the cube and an additional 0.5 for correctly placing it into the target container.

\paragraph{Simulated environments.}
We evaluate \ours in two established multi-task simulation benchmarks: LIBERO~\citep{liu2023libero} and Meta-World~\citep{yu2020metaworld}. LIBERO assesses diverse visuomotor skills across four categories--\textit{Spatial}, \textit{Object}, \textit{Goal}, and \textit{Long}--with 10 tasks per category (40 total). We use a dataset~\citep{kimopenvla,pertsch2025fast}\footnote{LIBERO dataset: \href{https://huggingface.co/datasets/physical-intelligence/libero}{physical-intelligence/libero}} containing 1,693 episodes covering all tasks, and evaluate with 10 trials per task, reporting average success rates based on binary completion criteria. 
Meta-World evaluates generalization across 50 tasks of varying difficulty: \textit{easy}, \textit{medium}, \textit{hard}, and \textit{very hard}~\citep{seo2023masked}. We use a dataset\footnote{Meta-World dataset: \href{https://huggingface.co/datasets/lerobot/metaworld_mt50}{lerobot/metaworld\_mt50}} of 2,500 episodes (50 per task), and mirror the evaluation protocol used for LIBERO: 10 trials per task, with trials scored as 1 only if the task is fully completed.

\begin{figure}
    \centering
    \begin{minipage}[t]{0.99\textwidth}
        \centering
        \includegraphics[width=\textwidth]{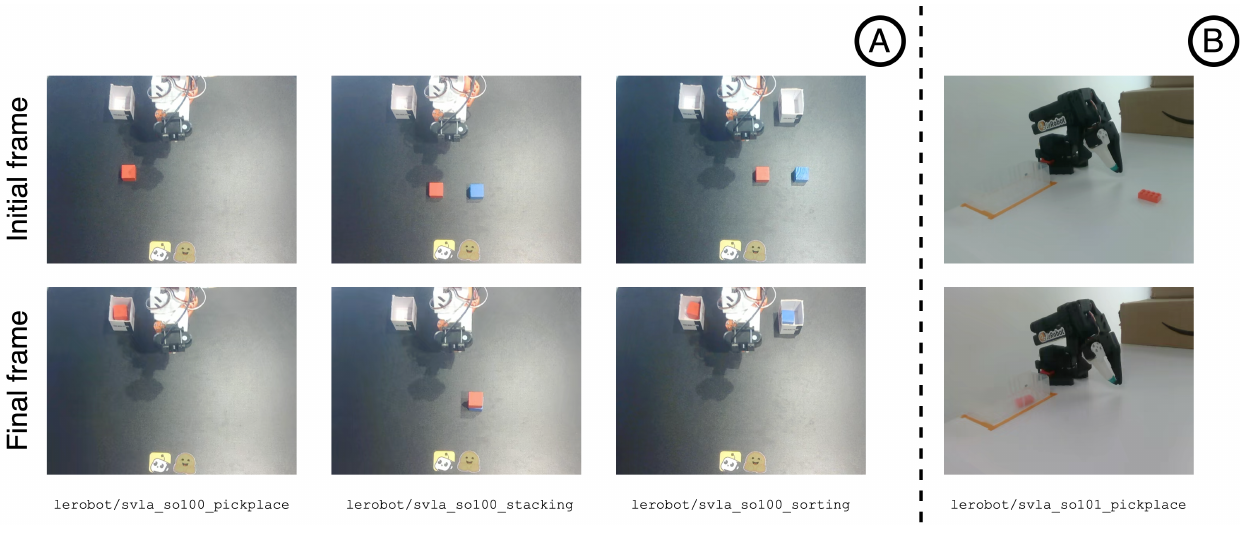}
        \caption{Illustrations of the four real-world tasks we benchmark \ours against, presenting starting and terminal frame for each of the dataset considered, for both SO100 embodiments (A) and SO101 (B). For SO100, we use top and wrist cameras, where for SO101 we use top and side cameras (as seen in the images).}
        \label{fig:tasks}
    \end{minipage}
    \vspace{-0.6cm}
\end{figure}

\paragraph{Real-world tasks.} We evaluated \ours on 4 datasets in a real-world setting, which we open-source on Hugging Face (\Cref{fig:tasks}). In particular, we benchmark real-world pick and placing capabilities\footnote{Pick-Place dataset: \href{https://huggingface.co/datasets/lerobot/svla_so100_pickplace}{lerobot/svla\_so100\_pickplace}}, stacking capabilities\footnote{Stacking dataset: \href{https://huggingface.co/datasets/lerobot/svla_so100_stacking}{lerobot/svla\_so100\_stacking}.}, and sorting capabilities\footnote{Sorting dataset: \href{https://huggingface.co/datasets/lerobot/svla_so100_sorting}{lerobot/svla\_so100\_sorting}.} for the SO100 robot, alongside real-world pick and placing capabilities for the SO101 platform \footnote{(SO101) Pick-Place dataset: \href{https://huggingface.co/datasets/lerobot/svla_so101_pickplace}{lerobot/svla\_so101\_pickplace}}. Critically, \ours is not pretrained on any datasets recorded for the SO101.

For the pick and place task, \ours is instructed to \texttt{pick up the cube and place it in the box}. The box is small in size and in a fixed position positions while the cube starting position is varied within 5 different starting conditions. %
We assess completion of the task with a fine-grained score resulting in a score of 0.5 for successfully grasping the cube, and 0.5 for successfully placing it into the box.
 
For the stacking task, \ours is required to put a cube on top of another. We instruct the robot to \texttt{pick up the red cube and put it on top of the blue cube}. The initial positions of both cubes vary across episodes.
We assess completion of the task with a fine-grained score resulting in a score of 0.5 for successfully grasping the top cube, and 0.5 for successfully placing it on top of the bottom cube.

For the sorting tasks, which has longer horizon, \ours must sort the cubes depending on the color, following the instruction to \texttt{put the red cube in the right box and the blue cube in the left box}. The cubes are placed in  5 different positions as in Task~1. To introduce variation, the colors of the cubes are flipped, with 5 episodes per color configuration, resulting in 10 demonstrations per position. The boxes locations remain fixed across all demonstrations. 
We assess completion of the task with a fine-grained score resulting in a score of 0.25 for successfully grasping either of the cubes, and 0.25 for successfully completing one cube-box matching, resulting in a score of 0.25 \( \times \) 4 upon task completion.
Figure~\ref{fig:tasks}(A) presents initial and final frames for successful episodes for all tasks, alongside the Hugging Face handle of the corresponding dataset\footnote{Datasets can be easily explored via \href{https://huggingface.co/spaces/lerobot/visualize_dataset}{\texttt{visualize\_dataset}}}.

To assess \ours's generalization, we also evaluate our model on a different robot embodiment and task\footnote{Pick-Place-Lego dataset: \href{https://huggingface.co/datasets/lerobot/svla_so101_pickplace}{lerobot/svla\_so101\_pickplace}.}, similar to pick-place but rather using a small block instead of a cube. 
In this task, the robot is instructed to \texttt{put the pink lego brick into the transparent box}. This task requires more precision, especially in grasping the small lego object, together with advanced vision capabilities considering the box's transparency.

\subsection{Robots}
Across simulation and real-world enviroments, we use a variety of robotic platforms.
\begin{itemize}
    \item \textbf{SO100 and SO101 \citep{cadene2024lerobot}}. The Standard Open SO-100 is a low-cost, 3D-printable robotic arm designed for improving accessibility to robotics and robot learning research. Both the SO-100 and its updated version, the SO-101, are open-source platforms for basic manipulation tasks. Each arm has six degrees of freedom and uses low-cost servo motors that are controlled with position commands. The SO101 has better arm design for faster assembly and different motors, making its movements smoother and better for tasks requiring more precisions.
    \item \textbf{Panda \citep{haddadin2022franka}.} The Franka Emika Panda is a single 7-DOF torque-controlled robotic arm designed for safe and precise manipulation. Its high-resolution joint sensing and compliant control make it well-suited for learning-based manipulation tasks in both simulation and real-world settings. This robot is used in the LIBERO simulator.
    \item \textbf{Swayer \citep{yu2020metaworld}.} Is a single 4-DOF controlled robotic arm designed for manipulation tasks. Is is used in the Meta-World simulator and the policy control the position and state of the gripper.    
\end{itemize}

\subsection{Implementation details.}

We conduct our experiments using LeRobot~\citep{cadene2024lerobot}, a PyTorch-based framework for real-world robotics. 
During pretraining, we train for 200,000 steps with a global batch size of 256 on all our community datasets. 
After a 100-step warmup, we use a cosine learning rate schedule starting at 1e-4 and decaying to a minimum of 2.5e-6. We use the AdamW optimizer with $\beta_1=0.9, \beta_2=0.95$.
Training is performed after resizing the images to 512×512, for consistency with the VLM input size. 
We use SmolVLM-2~\citep{marafioti2025smolvlm} as our VLM backbone. 
The action expert is trained with flow matching to output chunks of \( n = 50 \) actions. For real-world evaluation, we perform synchronous inference: the model samples new observations only after executing the full chunk of actions. In simulation, we perform inference by sampling new observations and predicting a new action after each executed action. During inference, the flow matching is fixed to 10 steps. 
We train only the action expert module, keeping the VLM frozen. 
Our main model, contains 450 million parameters, with approximately 100 million dedicated to the action expert. 
We use only the first 16 layers of the large language model (LLM) within the VLM. For fine-tuning on simulation benchmarks, we train for 100,000 steps with a batch size of 64, while for real-world tasks, we fine-tune for 200,000 steps. 
However, we observe in practice that the model can be trained for a much smaller number of steps without sacrificing significant performance levels.

Beyond maintaining a compact model and a reduced number of tokens, we employ several optimizations to enhance training efficiency. 
Specifically, we leverage \texttt{bfloat16} precision and \texttt{torch.compile()} \citep{paszke2019pytorch} that JIT-compiles PyTorch code into optimized kernels. 
To ensure compatibility with these optimizations, we maintain a fixed sequence length and batch size, discarding any excess frames in an episode that do not fit a complete batch. 
For multi-GPU and multi-node training, we utilize Hugging Face's \texttt{accelerate} \citep{accelerate} library with mixed precision, providing a scalable and memory-efficient training setup. 
Pretraining was conducted using 4 GPUs to accomodate for large batch size, but the model can easily be trained on a single GPU due to its small size. Overall, the project consumed approximately 30k GPU hours.

\subsection{Baselines}
We compare our model against two popular and strong baselines, both available in the LeRobot library \citep{cadene2024lerobot}.

\paragraph{\( \bf{\pi}_0 \) \citep{black2024pi_0}.}
\( \pi_0 \) is a VLA which leverages a VLM combined with Flow Matching for action chunk prediction. 
It has a total model size of 3.3 billion parameters and is pre-trained on 10,000 hours of cross-embodiment robotics data. 
The model architecture is based on Paligemma \citep{beyer2024paligemma} and accepts three RGB images, sensorimotor states, and a language instruction as inputs.

\paragraph{ACT \citep{zhao2023learningact}.}
ACT is a Conditional Variational Autoencoder (CVAE) \citep{NIPS2015_8d55a249cvae} policy model with an encoder-decoder transformer architecture containing approximately 80 million parameters. ACT uses a ResNet vision encoder pre-trained on ImageNet, while the CVAE is trained from scratch. The model generates action chunks and is optimized using a regression objective, directly predicting continuous actions. The model accepts a sequence of RGB images, and sensorimotor states.

\subsection{Main results}
In this section, we present the main results of \ours in both real-world and simulated environments. For real-world evaluation, \ours is pretrained on community-collected datasets. \( \pi_0 \) is finetuned on the respective target datasets, while ACT is trained from scratch on each dataset.

\begin{table}[h!]
    \centering
    \setlength{\tabcolsep}{6pt}
    \renewcommand{\arraystretch}{1.2}
    \definecolor{grayrow}{gray}{0.92}
    \resizebox{0.85\linewidth}{!}{
    \begin{tabular}{llc|ccccc}
        \toprule
        \textbf{Benchmark} & \textbf{Policy (\# Params)} & \textbf{VLA Pt.} & \multicolumn{5}{c}{\textbf{Success Rate (\%) — Simulation}} 
             \\
        \midrule
        \rowcolor{grayrow}
        \multicolumn{3}{l}{\textbf{LIBERO}} & \textbf{Spatial} & \textbf{Object} & \textbf{Goal} & \textbf{Long} & \textbf{Avg.}\\
        & Diffusion Policy \citep{khazatsky2024droid}  & No & 78.3 & 92.5 & 68.3 & 50.5 & 72.4 \\
        & Octo (0.09B) \citep{team2024octo} & Yes & 78.9 & 85.7 & 84.6 & 51.1 & 75.1 \\
        & OpenVLA (7B) \citep{kimopenvla} & Yes & 84.7 & 88.4 & 79.2 & 53.7 & 76.5 \\
        & $\pi_0$ (Paligemma-3B) & No & 87 & 63 & 89 & 48 & 71.8 \\
        & $\pi_0$ (3.3B) & Yes & 90 & 86 & 95 & 73 & 86.0 \\
        \midrule
        & \ours (0.24B) & No & 87 & 93 & 88 & 63 & 82.75 \\
        & \ours (0.45B) & No & 90 & 96 & 92 & 71 & 87.3 \\
        & \ours (2.25B) & No & 93	& 94 & 91 & 77 & \textbf{88.75} \\
        \midrule
        \rowcolor{grayrow}
        \multicolumn{3}{l}{\textbf{Meta-World}} & \textbf{Easy} & \textbf{Medium} & \textbf{Hard} & \textbf{Very Hard} & \textbf{Avg.} \\
        & Diffusion Policy \citep{chi2023diffusion} & No & 23.1 & 10.7 & 1.9 & 6.1 & 10.5 \\
        & TinyVLA \citep{TinyLLaVA} & No & 77.6 & 21.5 & 11.4 & 15.8 & 31.6 \\
        & $\pi_0$ (3.5B-Paligemma) & No & 80.4 & 40.9 & 36.7 & 44.0 & 50.5 \\
        & $\pi_0$ (3.5B) & Yes & 71.8 & 48.2 & 41.7 & 30.0 & 47.9 \\
        \midrule
        & \ours (0.24B) & No & 86.43 & 46.36 & 35 & 60 & 56.95 \\
        & \ours (0.45B) & No & 82.5 & 41.8 & 45.0 & 60.0 & 57.3 \\
        & \ours (2.25B) & No & 87.14 & 51.82 & 70 & 64 & \textbf{68.24} \\
        \bottomrule
    \end{tabular}
    }
    \caption{\textbf{Simulation benchmarks (LIBERO and Meta-World).} Success rates (\%) for various policies. LIBERO tasks correspond to different  settings (e.g., spatial, object, goal, long-horizon); Meta-World tasks reflect different difficulty levels. VLA Pt refers pretraining on robotics data. SmolVLA is only initialized from the VLM. Baselines scores from \cite{kimopenvla,wen2024tinyvla}.}
    \label{tab:libero_metaworld_combined}
\end{table}

\paragraph{Simulation Evaluation.}
In \Cref{tab:libero_metaworld_combined}, we further evaluate \ours on two major simulation benchmarks--LIBERO and Meta-World--using a multi-task training setup. \ours{} outperforms other VLA-based approaches such as Octo \citep{team2024octo} and OpenVLA \citep{kimopenvla}, as well as the diffusion policy baseline across both LIBERO and Meta-World. We also compare against two variants of $\pi_0$: one initialized from a vision-language model (Paligemma-3B), and another further pretrained on robotics datasets (intitialized from the weights released by the authors). Despite not being pretrained on robotics data, \ours{} consistently outperforms the VLM-initialized $\pi_0$ and performs competitively with the robotics-pretrained version. Note that, compared to $\pi_0$, \ours is around 40\% faster to train and consumes 6$\times$ less memory.

\begin{table}[h!]
    \centering
    \begin{minipage}[t]{0.52\linewidth}
        \centering
        \setlength{\tabcolsep}{10pt}
        \renewcommand{\arraystretch}{1.1}
        \definecolor{lightgray}{gray}{0.92}
        \resizebox{\linewidth}{!}{
        \begin{tabular}{lcccc}
            \toprule
            & \multicolumn{4}{c}{\textbf{Success Rate (\%) — Real World}} \\
            \cmidrule(lr){2-5}
            \textbf{Policy} & \textbf{Pick-Place} & \textbf{Stacking} & \textbf{Sorting} & \textbf{Avg.} \\
            \midrule
            \rowcolor{lightgray}
            \multicolumn{5}{l}{\textbf{Single-task Training}} \\
            ACT & 70 & 50 & 25 & 48.3 \\
            \midrule
            \rowcolor{lightgray}
            \multicolumn{5}{l}{\textbf{Multi-task Training}} \\
            $\pi_0$ (3.5B)       & 100 & 40 & 45 & 61.7 \\
            \ours (0.45B)         & 75 & 90 & 70 & 78.3 \\
            \bottomrule
        \end{tabular}
        }
        \caption{\textbf{Real-world benchmarks (SO100).} Success rate (\%) across three tasks using policies trained in multi-task and single-task settings.}
        \label{tab:main_results_real}
    \end{minipage}%
    \hspace{0.05\linewidth}%
    \begin{minipage}[t]{0.38\linewidth}
        \centering
        \setlength{\tabcolsep}{8pt}
        \renewcommand{\arraystretch}{1.2}
        \definecolor{lightgray}{gray}{0.92}
        \resizebox{\linewidth}{!}{
        \begin{tabular}{lcc}
            \toprule
            & \multicolumn{2}{c}{\textbf{Success Rate (\%) — Real World}} \\
            \cmidrule(lr){2-3}
            \textbf{Policy} & \textbf{In Distribution} & \textbf{Out of Distribution} \\
            \midrule
            \rowcolor{lightgray}
            \multicolumn{3}{l}{\textbf{Single-task Training}} \\
            ACT & 70 & 40 \\
            \ours (0.45B)  & 90 & 50 \\
            \bottomrule
        \end{tabular}
        }
        \caption{\textbf{Real-world benchmark (SO101).} Success rate (\%) for the Pick-Place-Lego task using policies trained in single-task setting.}
        \label{tab:main_results_real_so101}
    \end{minipage}
\end{table}

\paragraph{Real-World Evaluation.}
In \Cref{tab:main_results_real}, we evaluate \ours{} on four real-world tasks. For the SO101 benchmark, the model is trained on a combination of three datasets, and success rates are reported per task as well as on average. \ours{} outperforms both ACT \citep{zhao2023learningact}, which is trained individually on each task, and \( \pi_0 \), a significantly larger model in terms of parameter count (\( \sim 7\times\)). 
Similarly, on SO101 (see \Cref{tab:main_results_real_so101}), \ours{} surpasses ACT in both in-distribution and out-of-distribution (OOD) settings. For OOD evaluation, the Lego object is placed in novel positions not previously encountered during training.

\begin{table}[h!]
    \centering
    \begin{minipage}[t]{0.95\linewidth}
        \centering
        \setlength{\tabcolsep}{10pt}
        \renewcommand{\arraystretch}{1.1}
        \definecolor{lightgray}{gray}{0.92}
        \resizebox{0.7\linewidth}{!}{
        \begin{tabular}{lccccc}
            \toprule
            & & \multicolumn{4}{c}{\textbf{Success Rate (\%) — Real World}} \\
            \cmidrule(lr){3-6}
            \textbf{Policy} & \textbf{VLA pt.} &  \textbf{Pick-Place} & \textbf{Stacking} & \textbf{Sorting} & \textbf{Avg.} \\
            \midrule
            \rowcolor{lightgray}
            \multicolumn{6}{l}{\textbf{Single-task Training}} \\
            \ours (0.45B)   & No & 55 & 45 & 20 & 40 \\
            \midrule
            \rowcolor{lightgray}
            \multicolumn{6}{l}{\textbf{Multi-task Training}} \\
            \ours (0.45B)    & No   & 80 & 40 & 35 & 51.7 \\
            \ours (0.45B)    & Yes  & 75 & 90 & 70 & 78.3 \\
            \bottomrule
        \end{tabular}
        }
        \caption{\textbf{Effect of pretraining and multitask learning.} Success rate (\%) across three tasks using \ours trained in multi-task and single-task settings. Results with SO100 robot.}
        \label{tab:main_results_effect_of_pretrain}
    \end{minipage}%
    \hspace{0.05\linewidth}%

\end{table}

\paragraph{Effect of pretraining and multitask learning.} In \Cref{tab:main_results_effect_of_pretrain}, we further evaluate the impact of pretraining \ours on community datasets in terms of the difference in real-world performance, and investigate whether multitask finetuning provides additional benefits for \ours. The results show that, pretraining on community datasets leads to a substantial performance improvement (from 51.7 to 78.3). 
Furthermore, multitask finetuning yields further gains, underscoring the importance of knowledge transfer across tasks.

\begin{figure}[h!]
    \centering

    \begin{subfigure}[t]{0.42\linewidth}
        \centering
        \setlength{\tabcolsep}{5pt}
        \renewcommand{\arraystretch}{1.2}
        \resizebox{0.95\linewidth}{!}{
        \begin{tabular}{lcccc}
            \toprule
            \multirow{2}{*}{\textbf{Inference}} & \multicolumn{4}{c}{\textbf{Success Rate (\%) — Real World}} \\\cmidrule(lr){2-5}
             & \textbf{Pick-Place} & \textbf{Stacking} & \textbf{Sorting} & \textbf{Avg} \\
            \midrule
            Sync  & 75 & 90 & 70 & 78.3 \\
            Async & 80 & 90 & 50 & 73.3 \\
            \bottomrule
        \end{tabular}
        }
        \subcaption{\textbf{Performance (success rates).}}
        \label{tab:real_async_success}
    \end{subfigure}
    \hfill
    \begin{subfigure}[t]{0.28\linewidth}
        \centering
        \setlength{\tabcolsep}{5pt}
        \renewcommand{\arraystretch}{1.2}
        \resizebox{0.95\linewidth}{!}{
        \begin{tabular}{lccc}
            \toprule
            \multirow{2}{*}{\textbf{Inference}} & \multicolumn{3}{c}{\textbf{Time (s) — Real World}} \\\cmidrule(lr){2-4}
             & \textbf{Total} & \textbf{Avg} & \textbf{Std} \\
            \midrule
            Sync  & 137.5 & 13.75 & 2.42 \\
            Async & 97.0  & 9.70  & 2.95 \\
            \bottomrule
        \end{tabular}
        }
        \subcaption{\textbf{Task completion time.}}
        \label{tab:real_async_time}
    \end{subfigure}
    \hfill
    \begin{subfigure}[t]{0.28\linewidth}
        \centering
        \setlength{\tabcolsep}{5pt}
        \renewcommand{\arraystretch}{1.2}
        \resizebox{0.95\linewidth}{!}{
        \begin{tabular}{lccc}
            \toprule
            \multirow{2}{*}{\textbf{Inference}} & \multicolumn{3}{c}{\textbf{\# of Cubes — Real World} } \\\cmidrule(lr){2-4}
             & \textbf{Total} & \textbf{Avg} & \textbf{Std} \\
            \midrule
            Sync  & 9   & 1.8  & 0.45  \\
            Async & 19  & 3.8  & 1.3  \\
            \bottomrule
        \end{tabular}
        }
        \subcaption{\textbf{Performance in fixed time.}}
        \label{tab:real_async_cubes}
    \end{subfigure}

    \caption{\textbf{Comparison between synchronous (Sync) and asynchronous (Async) inference.} We evaluate \ours on three real-world tasks under both inference modes. Asynchronous inference achieves similar success rates (left) but is significantly faster (middle) and complete more tasks (right) in fixed-time settings. Hyperparameters have been optimized for Pick-Place and reused across tasks.}
    \label{fig:async_inference_summary}
\end{figure}

\subsection{Asynchronous inference}

We evaluated \ours{} under two inference modes: sync and async. 
The sync mode reflects a standard evaluation setting in robotics, whereby the policy predicts a chunk of actions that is executed fully before the next prediction cycle begins. In contrast, the async mode decouples action execution and policy inference, allowing predictions and control to run in parallel.

\paragraph{Results.}
For both inference modes, we report the success rate and policy speed (\Cref{fig:async_inference_summary}). To evaluate speed, we design two experiments using the Pick-Place task. In the first experiment, we measure the time taken to complete the task across 10 trials and 5 different cube positions. In the second, we fix a time limit (e.g., 60 seconds) and count the number of cubes successfully picked and placed into the box, from different positions. Timing begins when the robot starts moving. As shown in \Cref{tab:real_async_success}, both inference modes achieve comparable success rates across three real-world tasks. However, asynchronous inference demonstrates a substantial speed advantage (\Cref{tab:real_async_time}). On average, it completes the task in 9.7 seconds, compared to 13.75 seconds in the synchronous setting (\( \sim 30\% \) faster). Furthermore, under the fixed-time evaluation, the asynchronous mode allows the robot to complete 19 successful pick-and-place cycles, compared to only 9 in the synchronous case (\Cref{tab:real_async_cubes}). Qualitatively, we observe that asynchronous inference enables faster reactions and better adaptability to changes in the environment. The robot exhibits greater robustness to shifts in object positions and external disturbances, and overall is capable to solve the same tasks a significantly larger number of times due to the avoidance of prediction lags (\Cref{fig:async_inference_summary}).

\subsection{Ablation Study}
We conduct a comprehensive ablation study to assess key design choices behind the final \ours model. 
All ablations are conducted on the LIBERO benchmark. Unless otherwise noted, models are trained from scratch without any pretraining on robotics data. The VLM backbone is frozen, and only the action expert is trained from scratch.

\begin{figure}[h!]
    \centering
    \captionsetup{type=figure}
    
    \begin{minipage}[t]{0.48\linewidth}
        \centering
        \setlength{\tabcolsep}{6pt}
        \renewcommand{\arraystretch}{1.1}
        \resizebox{0.85\linewidth}{!}{
        \begin{tabular}{cccccc}
            \toprule
            \textbf{Attention} & \multicolumn{5}{c}{\textbf{Success Rate (\%) — LIBERO}} \\ \cmidrule(lr){2-6}
            \textbf{mechanism} & \textbf{S} & \textbf{O} & \textbf{G} & \textbf{10} & \textbf{Avg} \\
            \midrule
            CA             & 87 & 92 & 83 & 54 & 79.0 \\
            SA             & 80 & 94 & 84 & 40 & 74.5 \\
            CA+SA (ours)   & 86 & 99 & 90 & 67 & 85.5 \\
            \bottomrule
        \end{tabular}
        }
        \captionof{table}{\textbf{Cross vs self-attention.} Interleaved cross (CA) and self-attention (SA) yield the best results. 
        }
        \label{tab:ablation_model_attention}
    \end{minipage}
    \hfill
    \begin{minipage}[t]{0.48\linewidth}
        \centering
        \setlength{\tabcolsep}{6pt}
        \renewcommand{\arraystretch}{1.1}
        \resizebox{0.8\linewidth}{!}{
        \begin{tabular}{cccccc}
            \toprule
            \textbf{Attention} & \multicolumn{5}{c}{\textbf{Success Rate (\%) — LIBERO}} \\ \cmidrule(lr){2-6}
             \textbf{mask} & \textbf{S} & \textbf{O} & \textbf{G} & \textbf{10} & \textbf{Avg} \\
            \midrule
            Bidir   & 79 & 86 & 82 & 23 & 67.5 \\
            Causal  & 80 & 94 & 84 & 40 & 74.5 \\
            \bottomrule
        \end{tabular}
        }
        \captionof{table}{\textbf{Bidirectional vs causal attention.} Preventing future action leakage improves performance. 
        }
        \label{tab:ablation_causal_bidir_attention}
    \end{minipage}
\end{figure}

\paragraph{Cross-attention (CA) vs. self-attention (SA) between VLM and \( \actionexpert \).}
We compare how the VLM features interact with the action expert, comparing causal self-attention (SA), cross-attention (CA), or our proposed interleaved SA+CA setup. 
In the SA setting, action tokens attend to each other using a causal mask, while in the CA setting the VLM features act as keys and values for attention in the \( \actionexpert \). 
As shown in \Cref{tab:ablation_model_attention}, cross-attention outperforms self-attention significantly. Interleaving both yields the best results, highlighting their complementary strengths.

\begin{figure}[h!]
    \centering
    \captionsetup{type=figure}

    \begin{minipage}[t]{0.48\linewidth}
        \centering
        \setlength{\tabcolsep}{6pt}
        \renewcommand{\arraystretch}{1.2}
        \resizebox{0.8\linewidth}{!}{
        \begin{tabular}{cccccc}
            \toprule
            \multirow{2}{*}{\textbf{N}} & \multicolumn{5}{c}{\textbf{Success Rate (\%) — LIBERO}} \\\cmidrule(lr){2-6}
             & \textbf{S} & \textbf{O} & \textbf{G} & \textbf{10} & \textbf{Avg} \\
            \midrule
            8           & 77 & 88 & 86 & 49 & 75.0  \\
            16          & 88 & 91 & 91 & 44 & 78.5  \\
            24          & 86 & 97 & 86 & 49 & 79.5  \\
            32          & 89 & 94 & 85 & 53 & 80.3  \\
            \midrule
            Skip \%2     & 84 & 90 & 83 & 45 & 75.5  \\
            \midrule
            VLM-256M     & 86 & 83 & 75 & 59 & 75.8  \\
            \bottomrule
        \end{tabular}
        }
        \captionof{table}{\textbf{Skipping VLM layers.} Skipping layers from a large VLM (here 500M parameters) yields better results than downsizing the VLM. Skipping every second layer is a competitive baseline.}
        \label{tab:ablation_vlm_layers}
    \end{minipage}
    \hfill
    \begin{minipage}[t]{0.48\linewidth}
        \centering
        \setlength{\tabcolsep}{6pt}
        \renewcommand{\arraystretch}{1.2}
        \resizebox{0.8\linewidth}{!}{
        \begin{tabular}{cccccc}
            \toprule
            \textbf{Expert width} & \multicolumn{5}{c}{\textbf{Success Rate (\%) — LIBERO}} \\\cmidrule(lr){2-6}
             \textbf{(w.r.t. VLM)} & \textbf{S} & \textbf{O} & \textbf{G} & \textbf{10} & \textbf{Avg} \\
            \midrule
            $\times$1.00 & 87 & 96 & 90 & 56 & 82.3 \\
            $\times$0.75 & 82 & 89 & 84 & 55 & 77.5 \\
            $\times$0.50 & 89 & 94 & 85 & 53 & 80.3 \\
            $\times$0.25 & 76 & 97 & 83 & 39 & 73.8 \\
            \bottomrule
        \end{tabular}
        }
        \captionof{table}{\textbf{Expert capacity.} Adjusting the expert's hidden size affects performance. Larger capacities yield better success rates.}
        \label{tab:ablation_expert_capacity}
    \end{minipage}

\end{figure}

\paragraph{Causal vs. bidirectional attention on action tokens within \( \actionexpert \).} 
Next, we investigate how action tokens should attend to each other within the action expert, \( \actionexpert \). 
We compare: \emph{(i)} no interaction between action tokens (pure CA), \emph{(ii)} causal self-attention, and \emph{(iii)} bidirectional self-attention. \Cref{tab:ablation_causal_bidir_attention} shows causal self-attention performs best, while bidirectional interaction harms performance.
Surprisingly, the no-interaction (CA-only) setup performs competitively, suggesting that conditioning on VLM features alone can be powerful.

\paragraph{Using early LLM layers in the VLM.}
The VLM backbone consists of a vision encoder followed by an LLM. 
Motivated by improving the efficiency of \ours, we investigate using features from the first \( N < L\) layers only instead of all the \( L \) LLM layers available or the features~\citep{black2024pi_0}.
Before starting to train, we discard the top \( L - N \) layers of the VLM. 
As shown in \Cref{tab:ablation_vlm_layers}, using only the first half of the VLM layers gives a good trade-off between performance and compute. 
Further, we also test a variant sampling every second VLM layer (Skip \% 2, ~\citep{shukor2024skipping})--reducing depth by half while retaining full model capacity. \Cref{tab:ablation_vlm_layers} indicates skipping every second layer performs better than training a smaller VLM, but worse than using the first \( N < L \) layers directly.

\paragraph{Action Expert Capacity.}
Motivated by efficiency arguments, we investigate varying the hidden dimension of the action expert to explore the impact of model capacity on performance.
Given the VLM dimension \( d \), \Cref{tab:ablation_expert_capacity} shows reducing the expert's hidden size to \( 0.75 \times d \) achieves a good balance between performance and efficiency.

\begin{figure}[h!]
    \centering
    \captionsetup{type=figure}
    \begin{minipage}[t]{0.48\linewidth}
        \centering
        \setlength{\tabcolsep}{6pt}
        \renewcommand{\arraystretch}{1.2}
        \resizebox{0.9\linewidth}{!}{
            \begin{tabular}{cccccc}
            \toprule
                \textbf{Training} & \multicolumn{5}{c}{\textbf{Success Rate (\%) — LIBERO}} \\\cmidrule(lr){2-6}
                \textbf{objective} & \textbf{S} & \textbf{O} & \textbf{G} & \textbf{10} & \textbf{Avg} \\
                \midrule
                 Flow matching   & 89 & 94 & 85 & 53 & 80.25 \\
                 Regression   &  92	& 85 & 86 & 38 & 75.25 \\
            \bottomrule
            \end{tabular}%
            }
            \captionof{table}{\textbf{Training objective.} We compare our training objective based on Flow matching to regression with L1 loss.}
        \label{tab:ablation_regression}
    \end{minipage}
    \hfill
    \begin{minipage}[t]{0.48\linewidth}
        \centering
        \setlength{\tabcolsep}{6pt}
        \renewcommand{\arraystretch}{1.2}
        \resizebox{0.8\linewidth}{!}{
            \begin{tabular}{ccccccc}
                \toprule
                \textbf{States} & \textbf{Attention} & \multicolumn{5}{c}{\textbf{Success Rate (\%) — LIBERO}} \\\cmidrule(lr){3-7}
                 & & \textbf{S} & \textbf{O} & \textbf{G} & \textbf{10} & \textbf{Avg} \\
                \midrule
                Prefix & CA & 89 & 94 & 85 & 53 & 80.3 \\
                Suffix & CA & 86 & 82 & 78 & 47 & 73.3 \\
                Prefix & SA & 62 & 74 & 57 & 20 & 53.3 \\
                Suffix & SA & 80 & 92 & 80 & 47 & 74.8 \\
                \bottomrule
            \end{tabular}
        }
        \captionof{table}{\textbf{States as prefix vs. suffix.} Feeding states to the VLM (prefix) leads to better performance than feeding them to the expert (suffix).}
        \label{tab:ablation_states_to_prefix}
    \end{minipage}
\end{figure}

\paragraph{Regression vs. Flow Matching training objectives.}
We compare two learning objectives for training the action expert \( \actionexpert \): flow matching (our default), and a standard regression L1 loss on the predicted versus ground-truth action chunks. 
In keeping with~\citet{black2024pi_0, chi2024diffusionpolicy} \Cref{tab:ablation_regression} shows that flow matching significantly outperforms regression, suggesting flow matching provides better inductive bias for modeling complex, multimodal action distributions.

\paragraph{States to the VLM or Action Expert?}
We compare two variants: \emph{(i)} feeding the sensorimotor states to the VLM (by projecting them into token space), and \emph{(ii)} passing them directly to the action expert. 
\Cref{tab:ablation_states_to_prefix} indicates including state information in the VLM leads to significantly better performance for both the CA and SA variants.

\begin{figure}[h!]
    \centering
    \captionsetup{type=figure}
    
    \begin{minipage}[t]{0.48\linewidth}
        \centering
        \setlength{\tabcolsep}{6pt}
        \renewcommand{\arraystretch}{1.2}
        \resizebox{0.7\linewidth}{!}{
        \begin{tabular}{cccccc}
            \toprule
            \textbf{Chunk} & \multicolumn{5}{c}{\textbf{Success Rate (\%) — LIBERO}} \\\cmidrule(lr){2-6}
            \textbf{Size} & \textbf{S} & \textbf{O} & \textbf{G} & \textbf{10} & \textbf{Avg} \\
            \midrule
             1   & 45 & 77 & 54 & 24 & 50.0  \\
             10  & 90 & 94 & 94 & 58 & 84.0  \\
             30  & 85 & 94 & 87 & 48 & 78.5  \\
             50  & 89 & 94 & 85 & 53 & 80.3  \\
             100 & 83 & 88 & 85 & 42 & 74.5  \\
            \bottomrule
        \end{tabular}
        }
        \captionof{table}{\textbf{Action chunk size.} A chunk size between 10 and 50 strikes a good balance between action prediction frequency and performance.}
        \label{tab:ablation_chunk_size}
    \end{minipage}
    \hfill
    \begin{minipage}[t]{0.48\linewidth}
        \centering
        \setlength{\tabcolsep}{6pt}
        \renewcommand{\arraystretch}{1.2}
        \resizebox{0.7\linewidth}{!}{
        \begin{tabular}{cccccc}
            \toprule
            \textbf{Action} & \multicolumn{5}{c}{\textbf{Success Rate (\%) — LIBERO}} \\\cmidrule(lr){2-6}
             \textbf{Steps}& \textbf{S} & \textbf{O} & \textbf{G} & \textbf{10} & \textbf{Avg} \\
            \midrule
             1  & 89 & 94 & 85 & 53 & 80.3 \\
             10 & 89 & 94 & 91 & 57 & 82.8 \\
             30 & 76 & 91 & 74 & 42 & 70.8 \\
             50 & 54 & 70 & 58 & 25 & 51.8 \\
            \bottomrule
        \end{tabular}
        }
        \captionof{table}{\textbf{Action execution steps.} Sampling new observations more frequently (e.g., every 1 or 10 steps) significantly improves performance.}
        \label{tab:ablation_action_steps}
    \end{minipage}
\end{figure}

\paragraph{Action chunk size, \( n \).}
Our model predicts chunks of actions, where each chunk consists of \( n \) time steps. We study the effect of varying \( n \) on the overall performance. 
A larger \( n \) allows the robot to execute more actions at inference time before needing to process new observations and predict the next chunk. 
However, \Cref{tab:ablation_chunk_size} shows that both very small and very large values of \( n \) degrade performance. We find that chunk sizes between 10 and 50 provide a good balance between the robot reactivity and efficiency.

\paragraph{Number of executed actions before updating observations.}
To improve inference speed in real-world deployment, the robot can execute multiple actions from the predicted chunk before processing new observations, hereby overwriting the current chunk before its exhaustion.
Still, while acting the entire chunk speeds up inference, it also reduces the robot's responsiveness to environmental changes.
\Cref{tab:ablation_action_steps} demonstrates updating observations more frequently significantly improves success rate, highlighting a trade-off between inference speed and control accuracy.

\section{Discussion}

We introduce a compact, efficient, and lightweight  VLA model--\ours--that runs on consumer-grade hardware, controls low-cost robots, and rivals significantly larger VLAs. 
\ours's architecture is designed for efficient training and inference without compromising the success rate.
In addition, we propose an asynchronous inference stack that enables faster adaptation and responsiveness in real-world manipulation tasks. 
This inference strategy is model-agnostic and can be integrated with any policy that outputs chunks of actions. 
Our work is supported by thorough ablations and analysis of the proposed architecture, that can guide practitioners and researchers to further improve the model architecture. 
Finally, we open-source our model, codebase, training datasets, robots hardware and provide detailed instructions to facilitate full reproducibility.

\subsection{Limitations}
We identify several limitations remaining in our contribution. In particular:

\begin{itemize}
    \item{\textbf{Dataset diversity and cross-embodiment training.}} Our pretraining currently uses datasets collected from a single robot type (SO100). Although we demonstrate that the model can be fine-tuned to different robots (\Cref{tab:main_results_real_so101}) and outperforms existing baselines, we argue incorporating training data from multiple robot embodiments is likely to prove critical in enhancing the model’s ability to generalize to new robotic platforms.
    \item{\textbf{Dataset size and scalability.}} The dataset used for training contains approximately 23k trajectories, and is significantly smaller than those used in typical VLA training regimes--OpenVLA, for instance, utilizes around 1 million trajectories. Expanding the dataset size could substantially improve the model’s performance and its generalization across a wider range of tasks and environments.
    \item{\textbf{Model size and hardware efficiency.}}
    \ours has less than 0.5 billion parameters, allowing for fast inference on consumer-grade hardware. While this efficiency is beneficial, exploring ways to scale these architectures further without sacrificing speed or accessibility is an important direction for future research. 
    \item{\textbf{Choice of VLM backbone.}} We rely on an off-the-shelf VLM backbone pretrained mainly on document reading and OCR tasks~\citep{marafioti2025smolvlm}. However, it is not yet clear whether these VLMs are optimal for real-world robotic interaction scenarios. Future work could explore alternative or more specialized pretraining strategies to better align VLM backbones with the peculiar demands of robotic environments.
    \item{\textbf{Multimodal and robotics data joint training.}}
    Integrating shared training on both robotics-specific data and broader multimodal datasets has the potential to improve generalization and instruction-following abilities. Such joint training could lead to more robust and adaptable VLAs.
    \item{\textbf{Task complexity and longer horizon.}}
    While \ours competes effectively on relatively simple and short-horizon tasks, scaling the approach to tackle longer-horizon problems remains an important challenge. Incorporating hierarchical policies or multi-level planning mechanisms may help to address this complexity.
    \item{\textbf{Learning paradigms: Imitation vs. Reinforcement Learning.}} Our current approach primarily relies on imitation learning. Nevertheless, exploring reinforcement learning techniques for VLAs \citep{chen2025conrft}--especially for handling complex or long--horizon tasks—could offer significant performance benefits and more dexterous policy adaptation.
\end{itemize}

\section{Aknowledgements}
The authors thank Marina Barannikov, Alexandre Chapin, Ville Kuosmanen, and Jade Choghari for their help in community and simulated datasets. 
The authors thank Alexander Soare for his early work on asynchronous inference.
The authors thank Quentin Gallouedec, Pablo Montalvo-Leroux and the rest of the Hugging Face team for their support during this project. 
This work was partly supported by the HPC resources of IDRIS under the allocation 2025-[A0181016235] made by GENCI and the PostGenAI@Paris cluster (ANR-23-IACL-0007, FRANCE 2030).

\bibliographystyle{hfstyle/plainnat}
\bibliography{main}

\appendix

\section{Appendix}

\subsection{Community datasets}

\label{app:comm_data}

\paragraph{Task annotation}

For task annotation, we prompt the VLM with the following:
\begin{quote}
Here is a current task description: \{current\_task\}. Generate a very short, clear, and complete one-sentence describing the action performed by the robot arm (max 30 characters). Do not include unnecessary words. Be concise.\\
Here is some examples: Pick up the cube and place it in the box, open the drawer and so on.\\
Start directly with an action verb like ``Pick'', ``Place'', ``Open'', etc.\\
Similar to the provided examples, what is the main action done by the robot arm?
\end{quote}

\paragraph{List of datasets.}
\begin{verbatim}
satvikahuja/mixer_on_off_new_1, aergogo/so100_pick_place, andy309/so100_0314_fold_cloths
jchun/so100_pickplace_small_20250323_120056, astroyat/cube, Ofiroz91/so_100_cube2bowl
HappyPablo/dec3_data2, ZCM5115/so100_1210, francescocrivelli/orange_feeding
francescocrivelli/carrot_eating, 0x00raghu/toffee_red, 0x00raghu/toffee_red_2
0x00raghu/toffee_red_3__, 0x00raghu/toffee_blue, 0x00raghu/toffee_blue_2
0x00raghu/toffee_to_hand_1, 0x00raghu/toffee_to_hand_2, liyitenga/so100_bi_hello
liyitenga/so100_bi_giveme5, ZCM5115/so100_2Arm3cameras_movebox, pranavsaroha/so100_carrot_1
pranavsaroha/so100_carrot_3, pranavsaroha/so100_carrot_4, maximilienroberti/so100_lego_red_box
pranavsaroha/so100_squishy, rabhishek100/so100_train_dataset, pranavsaroha/so100_squishy100
swarajgosavi/kikobot_pusht_real_v2, pandaRQ/pickmed, swarajgosavi/act_kikobot_pusht_real
pranavsaroha/so100_squishy2colors, pranavsaroha/so100_squishy2colors_1, Chojins/chess_game_001_white
jmrog/so100_sweet_pick, Chojins/chess_game_002_white, pranavsaroha/so100_squishy2colors_2_new
Chojins/chess_game_003_white, aractingi/pick_place_lego_cube, Chojins/chess_game_004_white
Chojins/chess_game_005_white, Chojins/chess_game_006_white, Chojins/chess_game_007_white
koenvanwijk/blue2, jlitch/so100multicam3, koenvanwijk/blue52
jlitch/so100multicam6, aractingi/pick_place_lego_cube_1, jlitch/so100multicam7
vladfatu/so100_ds, Chojins/chess_game_000_white, HITHY/so100-kiwi
HITHY/so100_peach1, HITHY/so100_redstrawberry, satvikahuja/orange_mixer_1
satvikahuja/mixer_on_off, satvikahuja/orange_pick_place_new1, satvikahuja/mixer_on_off_new
danmac1/real_real332, FeiYjf/Makalu_push, liyitenga/so100_pick_taffy1
chmadran/so100_dataset04, FeiYjf/Maklu_dataset, FeiYjf/new_Dataset
liyitenga/so100_pick_taffy2, satvikahuja/mixer_on_off_new_4, CSCSXX/pick_place_cube_1.17
liyitenga/so100_pick_taffy3, liyitenga/so100_pick_taffy4, yuz1wan/so100_pick_pink
yuz1wan/so100_pick_wahaha, yuz1wan/so100_pp_pink, yuz1wan/so100_pour_cup
liyitenga/so100_pick_taffy5, liyitenga/so100_pick_taffy6, yuz1wan/so100_button
yuz1wan/so100_pickplace, liyitenga/so100_pick_taffy7, FeiYjf/push_gg
FeiYjf/push_0094, swarajgosavi/act_kikobot_block_real, liyitenga/so100_pick_taffy8
phospho-ai/OrangeBrick3Cameras, vaishanthr/toy_pick_place, SeanLMH/so100_picknplace_v2
pepijn223/yellow_lego_in_box1, DimiSch/so100_50ep_2, DimiSch/so100_50ep_3
SeanLMH/so100_picknplace, nbaron99/so100_pick_and_place2, chmadran/so100_dataset08
vaishanthr/toy_pickplace_50ep, Beegbrain/pick_place_green_block_lr, Ityl/so100_recording1
vaishanthr/toy_pickplace, ad330/so100_box_pickPlace, Beegbrain/so100_put_cube_cup
aractingi/push_green_cube_hf, aractingi/push_green_cube_hf_cropped_resized, 
carpit680/giraffe_task, carpit680/giraffe_sock_demo_1, DimiSch/so100_terra_50_2, 
carpit680/giraffe_sock_demo_2, aractingi/push_cube_to_face_reward 
aractingi/push_cube_to_face_reward_cropped_resized, aractingi/push_cube_reward_data
aractingi/push_cube_reward_data_cropped_resized, aractingi/push_cube_offline_data_cropped_resized, 
aractingi/push_cube_front_side_reward, aractingi/push_cube_front_side_reward_cropped_resized, 
aractingi/push_cube_front_side_reward_long, aractingi/push_cube_front_side_reward_long_cropped_resized
aractingi/push_cube_reward, aractingi/push_cube_reward_cropped_resized, 
aractingi/push_cube_square_reward_cropped_resized, aractingi/push_cube_square_reward_1, 
aractingi/push_cube_square_reward_1_cropped_resized, aractingi/push_cube_square_light_reward
aractingi/push_cube_square_light_offline_demo
aractingi/push_cube_square_light_offline_demo_cropped_resized
denghj/dataset_red_tape01, aractingi/push_cube_square_offline_demo, 
aractingi/push_cube_square_offline_demo_cropped_resized, Beegbrain/stack_two_cubes, FeiYjf/Test_NNNN
LegrandFrederic/Orange-brick-lower-resolution, aractingi/pick_place_lego_cube_cropped_resized
aractingi/push_cube_overfit, aractingi/push_cube_overfit_cropped_resized, HITHY/so100_peach
zaringleb/so100_cube_2, andreasBihlmaier/dual_arm_transfer_2025_02_16, zaringleb/so100_cube_4_binary
1g0rrr/reward_pickplace1, 1g0rrr/reward_pickplace1_cropped_resized, FeiYjf/Hold_Pieces
FeiYjf/Grab_Pieces, hegdearyandev/so100_eraser_cup_v1, jbraumann/so100_1902
liyitenga/so100_pick_taffy10, mikechambers/block_cup_5, zaringleb/so100_cube_5_linear
yuz1wan/so100_pickplace_0223_2, yuz1wan/so100_pickplace_0223_3, samsam0510/mj_data_temp
samsam0510/tape_insert_1, samsam0510/tape_insert_2, pengjunkun/so100_push_to_hole
Deason11/Random_Kitchen, 1g0rrr/reward_dataset_name2, 1g0rrr/reward_dataset_name2_cropped_resized
1g0rrr/offline_dataset_name2, 1g0rrr/offline_dataset_name2_cropped_resized
aractingi/push_cube_simp_cropped_resized, danielkr452/so100_work6
Loki0929/so100_100, yuz1wan/so100_fold_0227_1
yuz1wan/so100_fold_0227_2, speedyyoshi/so100_grasp_pink_block, lirislab/stack_two_red_cubes
lirislab/red_cube_into_mug, lirislab/green_lego_block_into_mug, lirislab/green_lego_block_into_mug_easy
kevin510/lerobot-cat-toy-placement, NONHUMAN-RESEARCH/SOARM100_TASK_VENDA_BOX, wangjl1512/pour_water
airthebear/so100_GL, zijian2022/noticehuman1, zijian2022/noticehuman2
kantine/so100_kapla_tower6, zijian2022/noticehuman5, zijian2022/llm40
Ashton3/lerobot-aloha, zijian2022/noticehuman50, AaronNewman/screwdriver_task_batch1
AaronNewman/screwdriver_task_batch2, AaronNewman/screwdriver_task_batch3, zijian2022/noticehuman60
zijian2022/noticehuman70, Bartm3/tape_to_bin, liuhuanjim013/so100_th_1
Pi-robot/barbecue_flip, Pi-robot/barbecue_put, wangjl1512/doll
sshh11/so100_orange_50ep_1, sshh11/so100_orange_50ep_2, DorayakiLin/so100_pick_cube_in_box
Bartm3/tape_to_bin2, luke250305/play_dice_250311.1, andy309/so100_0311_1152
sihyun77/suho_so100, sihyun77/si_so100, shreyasgite/so100_base_left
sihyun77/suho_red, liuhuanjim013/so100_block, andy309/so100_0313_no_wrist_camera
zijian2022/l9, zijian2022/n1_2, DorayakiLin/so100_stack_cube
andy309/so100_0313_no_wrist_camera_with_two_arms_cloths, joaoocruz00/so100_makeitD1, zijian2022/l10_1
zijian2022/l10_5, sihyun77/suho_red2, sihyun77/suho_angel
sihyun77/sihyun_king, acrampette/third_arm_01, Winster/so100_cube
1g0rrr/sam_openpi03, thedevansh/mar16_1336, hkphoooey/throw_stuffie
doujiangwang/task1_10epi_100000step, sihyun77/sihyun_3_17_1, acrampette/third_arm_02
imsyed00/so100_yellowbowl_pickplace_1, kumarhans/so100_tape_task, sihyun77/sihyun_main
doujiangwang/task2_10epi_100000step, kantine/industrial_robothon_buttons_expert
kantine/industrial_robothon_buttons_anomaly, kantine/industrial_robothon_hatchAndProbe_expert
kantine/industrial_robothon_hatchAndProbe_anomaly, Odog16/so100_tea_towel_folding_v1
zijian2022/so100_318, zijian2022/so100_318_1, 
Congying1112/so100_place_blue_bottle_with_two_cameras
Congying1112/so100_place_blue_bottle_with_two_cameras2, 
Congying1112/so100_place_blue_bottle_with_single_camera, pietroom/first_task_short
kantine/industrial_screws_sorting_expert, kantine/industrial_screws_sorting_anomaly, pietroom/second_task
zijian2022/c0, doujiangwang/task4_10epi_100000step, Congying1112/so100_switch_with_onhand_camera
HYAIYN/so100_get_orange_10epi, doujiangwang/task5_10epi_100000step, 1g0rrr/sam_openpi_cube_low10
1g0rrr/sam_openpi_cube_top10, 1g0rrr/sam_openpi_wire10, 1g0rrr/sam_openpi_solder1
1g0rrr/sam_openpi_solder2, wcode/so100_put_pen_50, jchun/so100_pickplace_small_20250322_193929
bnarin/so100_tic_tac_toe_we_do_it_live, dc2ac/so100-t5, chmadran/so100_home_dataset
baladhurgesh97/so100_final_picking_3, bnarin/so100_tic_tac_toe_move_0_0
bnarin/so100_tic_tac_toe_move_1_0, bnarin/so100_tic_tac_toe_move_2_1
bnarin/so100_tic_tac_toe_move_4_0, zaringleb/so100_cube_6_2d
andlyu/so100_indoor_0, andlyu/so100_indoor_2, Winster/so100_sim
badwolf256/so100_twin_cam_duck, Congying1112/so100_simplepick_with_2_cameras_from_top, 
andlyu/so100_indoor_4, Zak-Y/so100_grap_dataset
kantine/domotic_pouringCoffee_expert, kantine/domotic_pouringCoffee_anomaly
lucasngoo/so100_strawberry_grape, kantine/domotic_makingCoffee_expert
kantine/domotic_makingCoffee_anomaly, ZGGZZG/so100_drop1
kantine/industrial_soldering_expert, kantine/industrial_soldering_anomaly
Yotofu/so100_sweeper_shoes, kantine/domotic_dishTidyUp_expert
kantine/domotic_dishTidyUp_anomaly, kantine/domotic_groceriesSorting_expert
kantine/domotic_groceriesSorting_anomaly, badwolf256/so100_twin_cam_duck_v2
kantine/domotic_vegetagblesAndFruitsSorting_expert
kantine/domotic_vegetagblesAndFruitsSorting_anomaly, kantine/domotic_setTheTable_expert
kantine/domotic_setTheTable_anomaly, therarelab/so100_pick_place, abhisb/so100_51_ep
andlyu/so100_indoor_val_0, lizi178119985/so100_jia, badwolf256/so100_twin_cam_duck_v3
andrewcole712/so100_tape_bin_place, Gano007/so100_lolo, Zak-Y/so100_three_cameras_dataset
Gano007/so100_doliprane, XXRRSSRR/so100_v3_num_episodes_50, zijian2022/assemblyarm2
ganker5/so100_action_20250403, andlyu/so100_indoor_val2, Gano007/so100_gano
paszea/so100_whale_grab, paszea/so100_whale, Clementppr/lerobot_pick_and_place_dataset_world_model
andlyu/so100_indoor_10, RasmusP/so100_dataset50ep_a, RasmusP/so100_dataset50ep
Gano007/so100_second, zaringleb/so100_cude_linear_and_2d_comb, dsfsg/grasp_pens
zijian2022/digitalfix, zijian2022/digitalfix2, zijian2022/digitalfix3
T1g3rGE/so100_pickplace_small_20250407_171912, sihyun77/mond_13
abokinala/sputnik_100_11_pick_place_container, dsfsg/bring_bottle
abokinala/sputnik_100_12_pick_place_container, Mwuqiu/so100_0408
AK51/4090_01, 356c/so100_rope_reposition_1, paszea/so100_lego_mix
abokinala/sputnik_100_14_pick_place_container, abokinala/sputnik_100_23_pick_place_surface, 
jiajun001/eraser00_2, jlesein/TestBoulon2 
duthvik/sputnik_100_31_pour_liquid, duthvik/sputnik_100_24_pick_place_surface
duthvik/sputnik_100_25_pick_place_surface, duthvik/sputnik_100_17_pick_place_container
duthvik/sputnik_100_26_pick_place_surface, VoicAndrei/so100_banana_to_plate_rebel_full
isadev/bougies1, danaaubakirova/so100_task_1
danaaubakirova/so100_task_2, danaaubakirova/so100_task_3, danaaubakirova/so100_task_4
sixpigs1/so100_pick_cube_in_box_error, sixpigs1/so100_push_cube_error, sixpigs1/so100_pull_cube_error
isadev/bougies2, therarelab/med_dis_rare_6, duthvik/sputnik_100_27_pick_place_surface
zijian2022/closer3, duthvik/sputnik_100_41_custom_tasks, duthvik/sputnik_100_42_custom_tasks
duthvik/sputnik_100_43_custom_tasks, duthvik/sputnik_100_44_custom_tasks, 
duthvik/sputnik_100_51_kitchen_tasks, duthvik/sputnik_100_52_kitchen_tasks
duthvik/sputnik_100_53_kitchen_tasks, duthvik/sputnik_100_45_custom_tasks
duthvik/sputnik_100_32_pour_liquid, duthvik/sputnik_100_29_pick_place_surface
duthvik/sputnik_100_18_pick_place_container
sixpigs1/so100_pull_cube_by_tool_error, sixpigs1/so100_insert_cylinder_error
abokinala/sputnik_100_54_kitchen_tasks, abokinala/sputnik_100_55_kitchen_tasks, m1b/so100_bluelego
abokinala/sputnik_100_46_custom_tasks
m1b/so100_bluelego_updt, kantine/flip_A0, kantine/flip_A1
kantine/flip_A2, kantine/flip_A3, lirislab/guess_who_no_cond
kantine/flip_A4, kantine/flip_A5, lirislab/guess_who_lighting
nguyen-v/so100_press_red_button, nguyen-v/so100_bimanual_grab_lemon_put_in_box2, pierfabre/cow
nguyen-v/press_red_button_new, nguyen-v/so100_rotate_red_button, Cidoyi/so100_all_notes
roboticshack/team10-red-block, Cidoyi/so100_all_notes_1, roboticshack/team_5-QuiEstCe_everyBox
roboticshack/team11_pianobot, roboticshack/team2-guess_who_so100
roboticshack/team2-guess_who_so100_light, roboticshack/team2-guess_who_so100_edge_case
roboticshack/team2-guess_who_less_ligth, Cidoyi/so100_all_notes_3
dsfsg/grasp_pen_and_bottle, abokinala/sputnik_100_60_kitchen_tasks
abokinala/sputnik_100_58_kitchen_tasks, danaaubakirova/so100_v2_task_1
danaaubakirova/so100_v2_task_2, danaaubakirova/so100_v2_task_3
danaaubakirova/so100_v2_task_4, zijian2022/force1, zijian2022/force2
zijian2022/force3, jiajun001/eraser00_3, zijian2022/bi2
zijian2022/bi1, zijian2022/hand1, Setchii/so100_grab_ball
MossProphet/so100_square-1-2-3.2, pierfabre/rabbit
bensprenger/right_arm_p_brick_in_box_with_y_noise_v0
pierfabre/horse, pierfabre/pig2, pierfabre/pig3
pierfabre/cow2, pierfabre/sheep, Chojins/chess_game_009_white
sihyun77/suho_3_17_1, sihyun77/sihyun_3_17_2, sihyun77/suho_3_17_3
sihyun77/sihyun_3_17_5, Odog16/so100_cube_drop_pick_v1, sihyun77/sihyun_main_2
sihyun77/suho_main_2, Bartm3/dice2, sihyun77/sihyun_main_3
Loki0929/so100_duck, pietroom/holdthis, pietroom/actualeasytask
Beegbrain/pick_lemon_and_drop_in_bowl, Beegbrain/sweep_tissue_cube, zijian2022/321
gxy1111/so100_pick_place, Odog16/so100_cube_stacking_v1, sihyun77/mond_1
andlyu/so100_indoor_1, andlyu/so100_indoor_3, frk2/so100large
lirislab/sweep_tissue_cube, lirislab/lemon_into_bowl, lirislab/red_cube_into_green_lego_block
lirislab/red_cube_into_blue_cube, 00ri/so100_battery, frk2/so100largediffcam
FsqZ/so100_1, ZGGZZG/so100_drop0, Chojins/chess_game_000_white_red
smanni/train_so100_fluffy_box, ganker5/so100_push_20250328, ganker5/so100_dataline_0328
ganker5/so100_color_0328, CrazyYhang/A1234-B-C_mvA2B, RasmusP/so100_Orange2Green
sixpigs1/so100_pick_cube_in_box, ganker5/so100_push_20250331, ganker5/so100_dataline_20250331
lirislab/put_caps_into_teabox, lirislab/close_top_drawer_teabox, lirislab/open_top_drawer_teabox
lirislab/unfold_bottom_right, lirislab/push_cup_target, lirislab/put_banana_bowl
Chojins/chess_game_001_blue_stereo, Chojins/chess_game_001_red_stereo, ganker5/so100_toy_20250402
Gano007/so100_medic, 00ri/so100_battery_bin_center, paszea/so100_whale_2
lirislab/fold_bottom_right, lirislab/put_coffee_cap_teabox, therarelab/so100_pick_place_2
paszea/so100_whale_3, paszea/so100_whale_4, paszea/so100_lego
LemonadeDai/so100_coca, zijian2022/backgrounda, zijian2022/backgroundb
356c/so100_nut_sort_1, Mwuqiu/so100_0408_muti, aimihat/so100_tape
lirislab/so100_demo, 356c/so100_duck_reposition_1, zijian2022/sort1
weiye11/so100_410_zwy, VoicAndrei/so100_banana_to_plate_only, sixpigs1/so100_stack_cube_error
isadev/bougies3, zijian2022/close3, bensprenger/left_arm_yellow_brick_in_box_v0
lirislab/guess_who_so100, bensprenger/left_arm_yellow_brick_in_box_with_purple_noise_v0, 
roboticshack/team16-can-stacking, zijian2022/insert2
roboticshack/team-7-right-arm-grasp-tape, Jiangeng/so100_413
roboticshack/team9-pick_cube_place_static_plate
AndrejOrsula/lerobot_double_ball_stacking_random 
roboticshack/left-arm-grasp-lego-brick
roboticshack/team-7-left-arm-grasp-motor, roboticshack/team9-pick_chicken_place_plate
roboticshack/team13-two-balls-stacking, tkc79/so100_lego_box_1
roboticshack/team13-three-balls-stacking, pierfabre/chicken
roboticshack/team16-water-pouring, ad330/cubePlace, Jiafei1224/so100_pa222per
paszea/so100_lego_2cam, bensprenger/chess_game_001_blue_stereo, Mohamedal/put_banana
tkc79/so100_lego_box_2, samanthalhy/so100_herding_1, jlesein/TestBoulon7
pranavsaroha/so100_onelego2, pranavsaroha/so100_onelego3, pranavsaroha/so100_carrot_2
vladfatu/so100_above, koenvanwijk/orange50-1, CSCSXX/pick_place_cube_1.18
dragon-95/so100_sorting, dragon-95/so100_sorting_1, nbaron99/so100_pick_and_place4
Beegbrain/pick_place_green_block, dragon-95/so100_sorting_3, HITHY/so100_peach3
shreyasgite/so100_legocube_50, triton7777/so100_dataset_mix, NONHUMAN-RESEARCH/SOARM100_TASK_VENDA
mikechambers/block_cup_14, samsam0510/tooth_extraction_3, samsam0510/tooth_extraction_4
samsam0510/cube_reorientation_2, samsam0510/cube_reorientation_4, samsam0510/glove_reorientation_1
vladfatu/so100_office, pranavsaroha/so100_legos4, Ityl/so100_recording2
FeiYjf/new_GtoR, dragon-95/so100_sorting_2, HITHY/so100_peach4
jpata/so100_pick_place_tangerine, HITHY/so100_strawberry, shreyasgite/so100_base_env
koenvanwijk/orange50-variation-2, pranavsaroha/so100_carrot_5, pandaRQ/pick_med_1
aractingi/push_cube_offline_data, DorayakiLin/so100_pick_charger_on_tissue, zijian2022/noticehuman3
liuhuanjim013/so100_th
\end{verbatim}
\end{document}